\documentclass[twocolumn, switch]{article}
\usepackage{preprint}
\usepackage{hyperref}
\usepackage{amsmath}
\usepackage{tabularx}
\usepackage{booktabs}
\usepackage{caption}
\usepackage[numbers,square]{natbib}
\usepackage{url}      
\usepackage{breakurl} 

\hypersetup{colorlinks=true, linkcolor=purple, urlcolor=blue, citecolor=cyan, anchorcolor=black}

\usepackage[utf8]{inputenc}	
\usepackage[T1]{fontenc}
\usepackage{xcolor}
\usepackage{lineno}					
\usepackage{tikz} 					

\usepackage{newfloat}
\DeclareFloatingEnvironment[name={Supplementary Figure}]{suppfigure}
\usepackage{sidecap}
\sidecaptionvpos{figure}{c}

\usepackage{titlesec}
\titlespacing\section{0pt}{12pt plus 3pt minus 3pt}{1pt plus 1pt minus 1pt}
\titlespacing\subsection{0pt}{10pt plus 3pt minus 3pt}{1pt plus 1pt minus 1pt}
\titlespacing\subsubsection{0pt}{8pt plus 3pt minus 3pt}{1pt plus 1pt minus 1pt}

\definecolor{lime}{HTML}{A6CE39}

\title{A Self-Supervised Framework for Space Object Behaviour Characterisation}

\usepackage{authblk}

\author[1, *]{Ian Groves}
\author[2]{Andrew Campbell}
\author[3]{James Fernandes}
\author[3]{Diego Ramírez Rodríguez}
\author[2]{Paul Murray}
\author[4]{Massimiliano Vasile}
\author[1]{Victoria Nockles}

\affil[1]{Defence AI Research Centre, Defence \& National Security, The Alan Turing Institute, UK}

\affil[2]{Department of Electronic and Electrical Engineering, University of Strathclyde, UK}
\affil[3]{GMV, UK}
\affil[4]{Aerospace Centre of Excellence, University of Strathclyde, UK}
\affil[*]{Corresponding author: \texttt{igroves@turing.ac.uk}}

\begin{document}

\twocolumn[\begin{@twocolumnfalse}
\maketitle
\begin{abstract}
Foundation Models, which leverage large neural networks pre-trained on unlabelled data before fine-tuning for specific tasks, are increasingly being applied to specialised domains. Recent examples include ClimaX for climate and Clay for satellite Earth observation, but a Foundation Model for Space Object Behavioural Analysis (SOBA) has not yet been developed. As orbital populations grow, automated methods for characterising space object behaviour are crucial for space safety. Here, we present a self-supervised framework for SOBA, representing a first step towards a Foundation Model for Space Object Behavioural Analysis.
The backbone of this framework is a Perceiver-Variational Autoencoder (VAE) architecture, pre-trained with self-supervised reconstruction and masked reconstruction on \textasciitilde 227,000 light curves from the MMT-9 observatory. The VAE enables anomaly detection, space object motion prediction, and generation of synthetic light curves. We fine-tuned the model for anomaly detection \& motion prediction using two independent light curve simulators (CASSANDRA and GRIAL respectively), using CAD models of boxwing, Sentinel-3, SMOS, and Starlink platforms.
Our pre-trained model achieved a reconstruction mean squared error of 0.0012, identifying potentially anomalous light curves through reconstruction difficulty. After fine-tuning, the model scored 85\% accuracy (0.92 ROC AUC) on anomaly detection and 82\% accuracy (0.95 ROC AUC) on motion mode prediction (e.g., sun-pointing, spin, tumbling). Analysis of high-confidence anomaly predictions on real data revealed distinct patterns including characteristic object profiles and satellite glinting.
The motion prediction model successfully differentiated between various movement behaviours such as sun-pointing, spin, and tumbling.
Our work demonstrates how self-supervised learning can simultaneously enable anomaly detection, motion prediction, and synthetic data generation from rich representations learned in pre-training.
More broadly, our work supports space safety and sustainability through automated monitoring and simulation capabilities.
\end{abstract}
\vspace{0.5cm}

\vspace{0.5cm}
\end{@twocolumnfalse}]


\section{Introduction}\label{sec1}
\begin{figure*}[t]
\centerline{\includegraphics[width=0.9\textwidth]{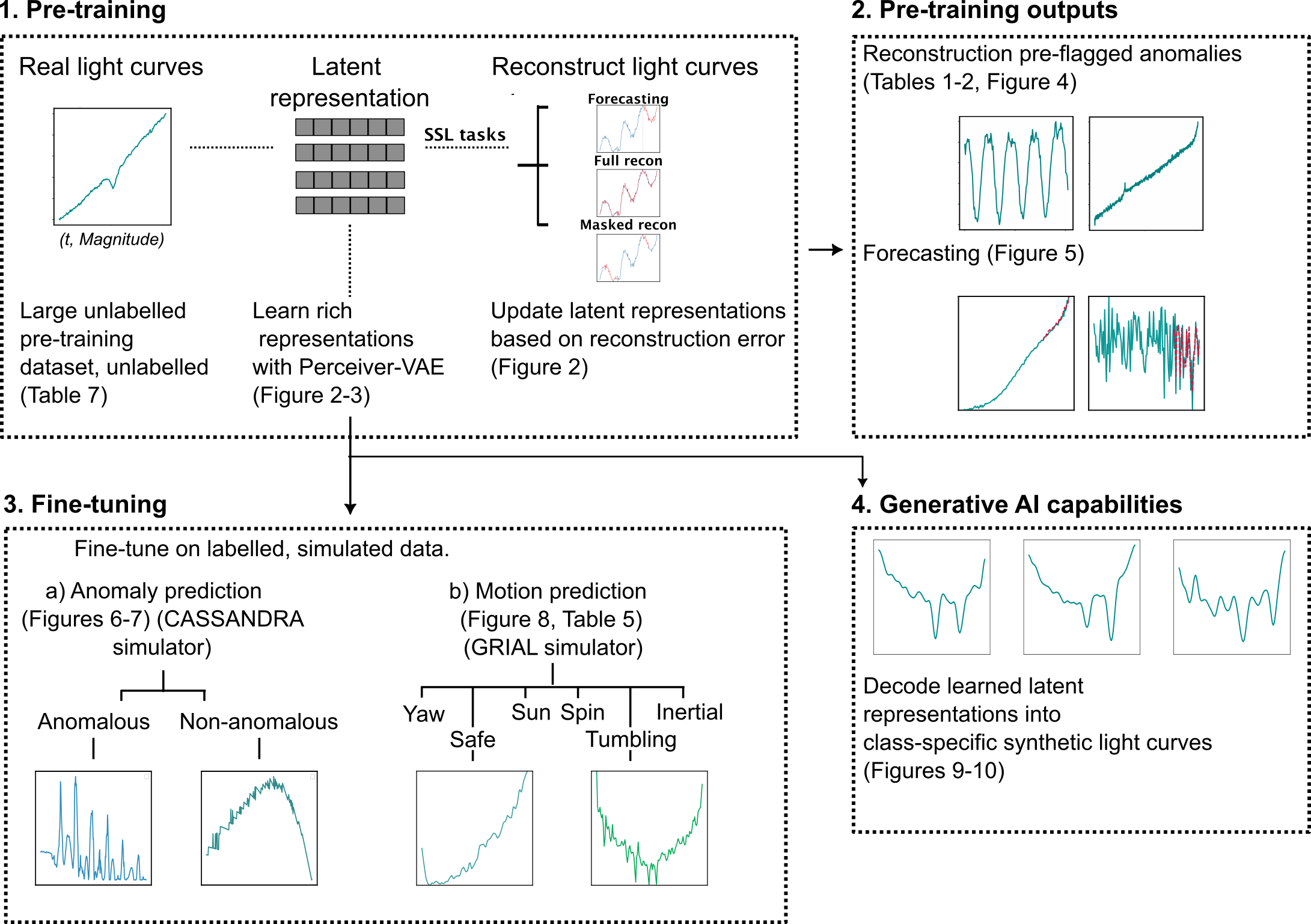}}
\caption{Graphical structure of the paper. First, in Section \ref{results:pre-training-training} we describe pre-training our model with a large unlabelled dataset of real light curves. We encode these into a rich latent representation with a Perceiver-VAE architecture, updating these representations based on three self-supervised learning (SSL) tasks: Reconstruction, Forecasting, and Masking. 
We next analyse the results of pre-training, which includes pre-flagging of anomalies based on reconstruction difficulty (Sections \ref{sec:pretraining_preflagging}-\ref{sec:pre-flagged_anomalies_analysis}), and forecasting quality of the model (\ref{sec:forecasting}).
Following this, we describe fine-tuning our rich representations for two downstream tasks: a) anomaly prediction (Section \ref{sec:anomaly_detection_fine-tuning}), and b) motion prediction (Section \ref{sec:fine-tuning_motion}).
Finally, we demonstrate further utility of our representations by generating de novo datasets according to a particular motion type (Section \ref{sec:synthetic_data_generation}).
}
\label{fig:graphical_abstract}
\end{figure*}
The number of objects launched is growing rapidly, with 159 worldwide launches in 2000 versus 2849 in 2024 \citep{UNSpaceLaunches}.
This growth underscores the need for methods for automated and efficient monitoring of space objects, which is crucial to societal function and national security e.g., critical communications \& position/navigation/timing \citep{npsa_critical_nodate}.
In addition to national security, monitoring of Space Objects has substantial economic  considerations, e.g., the financial service sector is reliant on precise time synchronisation enabled by satellites \cite{lombardi_accurate_2016}.
This increase in space resident objects is coupled with significantly larger datasets collected from modern sensors, tracking objects both from ground-based sensors and space-based sensors \citep{maxar_2019, HEO2025}.
Traditional methods for Space Object Behavioural Analysis (SOBA) rely on arduous manual inspection, or numerical methods requiring a large amount of priors. 
Machine learning and Artificial Intelligence techniques offer promising new possibilities to analyse these growing datasets. For example, Foundation Models (FMs), whereby large neural networks are (pre-)trained to learn general principles of unlabelled data, are emerging as a powerful approach for pattern recognition in large, unstructured datasets. They have demonstrated strong performance in multiple natural language and text-based tasks, e.g., writing and coding.
However, Foundation Models that integrate sensor data with physics-based models, are still in their infancy, as there are fewer publicly available datasets and input/output data types are less intuitive than natural language. 
Nevertheless, domain-specific Foundation Models are proving effective in specialised use-cases such as ClimaX \citep{nguyen_climax_2023} for climate/weather prediction, and Clay for Earth Observational data \citep{noauthor_clay_nodate}. 
These examples show that the general principles of learning spatiotemporal relationships between datapoints are  readily transferable and useful outside of natural language and programming. 

However, there is little current exploration on the best neural network architectures and training strategies for FMs for more specialised datasets/applications. To our knowledge, this work represents the first step towards a Foundation Model for Space Object Behavioural Analysis. Whilst our model is pre-trained on a single modality (light curves), the Perceiver architecture is well suited 
to multimodal extension, and we discuss this direction in Section~\ref{outlook}. We therefore use the term Foundation Model as a framing for this approach, following the precedent of domain-specific models such as ClimaX and Clay.
Of principal interest in Space Domain Awareness (SDA) is anomaly detection, i.e., detecting space objects that are behaving in an unusual way, such as motions, manoeuvres, or those that are performing atypical functions. 
Characterising these atypical behaviours can be done through simulation of space object observations, but this is challenging, due to imperfect physical models and computational cost.
Therefore, general purpose models for SDA should be able to perform anomaly detection, integrating real observations with physical models in a computationally efficient way.
\begin{figure*}[t!]
\centerline{\includegraphics[width=\textwidth]{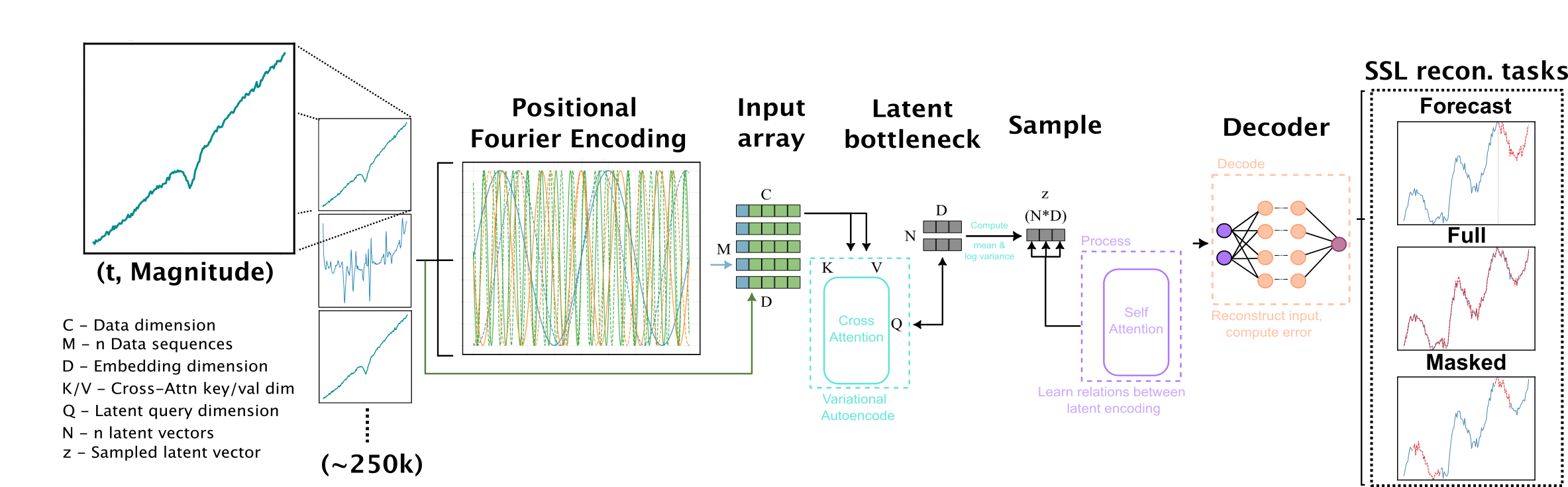}}
\caption{Pre-training approach. Input array(s) (M) provide keys (K) which index the data (e.g., timestep in a timecourse) and values (V) which represent the information at each K. The model includes a learned latent array (N) which provides queries (Q) for the Cross Attention mechanism. Q interacts with K and V to extract relevant information from the input. The latent representation (z) is computed from this mechanism using the mean and log variance, capturing compressed features of the input. In this way, z acts as a latent bottleneck. Self Attention layers then learn meaningful relations within this latent space. This architecture incorporates elements of Variational Autoencoders (VAE) in its training process, whereby the loss is calculated by sampling probabilistically from the latent space. Figure adapted from \citep{jaegle_perceiver_2021}.}
\label{fig:architecture-pretraining}
\end{figure*}
Foundation Models are typically pre-trained on large datasets to learn some compressed rich, general latent representation of the data distribution. They are then fine-tuned for specific tasks by further using those latent representations, for example, classifying them according to a small amount of labelled data.
Whilst in pre-training, the goal is usually to keep the optimisation task relatively unconstrained to encourage general features to be learned, we can still make architectural decisions to build in anomaly detection from the ground up through the use of Variational Autoencoder (VAE) components.

VAEs are generative models which are trained by learning to reconstruct the input data from a sampling of the learned latent space \citep{kingma_auto-encoding_2022}. 
They provide a simple and intuitive way to perform anomaly detection, as they inherently produce a reconstruction error at inference. 
This error is the difference between the model’s reconstruction of an input and the
actual input. 
If a specific data point is unusual compared to the training data, or is not well represented in the training data, it will produce a high reconstruction error, suggesting anomalous behaviour or properties. 
Unsupervised models with Variational Autoencoders (VAEs) have been used extensively as anomaly detectors, (for a recent review in the context of solar images see the Introduction in \cite{giger_unsupervised_2024}). 

Whilst VAEs offer an effective anomaly detection mechanism, they still need integration into a broader neural network framework to effectively process complex, high dimensional space object data.
Transformer architectures are an attractive choice here, due to their ability to capture long range dependencies in sequential data.
When choosing a Transformer architecture, there are numerous variants with respective strengths and weaknesses.
For example, conventional transformer-based neural networks perform the computationally costly full self attention mechanism between all the inputs. This means that for very long sequences, standard transformers incur quadratic computational cost with respect to sequence length, which can become prohibitive at scale.
To address this limitation, DeepMind recently developed the Perceiver/Perceiver-IO, a transformer-based neural network architecture \citep{jaegle_perceiver_2021, jaegle_perceiver_2022}.
Whilst a light-curve specific model may benefit from a state-of-the-art 
architecture for time-series analysis such as TimesNet~\citep{wu_timesnet_2023} 
for anomaly detection or TimeXer~\citep{wang_timexer_2024} for forecasting, 
for our framework, which aims to provide a first step towards an SDA Foundation Model, we prioritise extensibility to 
other observational modalities (e.g., hyperspectral curves, ISAR images, 
attitude measurements, orbital parameters). To achieve this multimodal 
compatibility, we adopt the Perceiver architecture, which is straightforwardly 
extendable to multiple modalities through its cross-attention latent bottleneck. 
This choice also reduces the computational complexity of the vanilla transformer from
$\mathcal{O}(N^2)$ to $\mathcal{O}(N)$ with respect to sequence length, an 
advantage that is primarily forward-looking in this work but could support 
edge deployment (e.g., on-board a satellite).

Given the potential of VAEs as anomaly detectors, and the computational efficiency
of Perceiver, a VAE-based Perceiver model is a promising approach for a computationally efficient, SDA FM for anomaly detection.
In this work, we use visible light curves as a test-bed for our self-supervised framework, providing proof of principle for an SDA FM. Light curves are readily available in large volumes of real observational data, and light curve simulators are relatively mature. Because of this, we use the largest publicly released dataset of light curves, from the Mini-Mega TORTORA (MMT-9) observatory in Russia \citep{beskin_wide-field_2017}. 

Light curves are plotted timecourse measurements of the brightness of light received by a sensor reflected by, or emitted from a space object (SO). Light curves allow both the physical properties and the behaviour of the SO to be inferred. 
For example, rotating objects may exhibit short-term periodic variations in their light curves. 

Light curves are well studied in the literature and recently there has been substantial interest in the field in using light curves to train machine learning models \citep{bertolini_space_2022, furfaro_space_2018, qashoa_classification_2023}.
These efforts are mostly concerned with supervised classification problems; whereby light curves are labelled into classes, and the model is trained to distinguish between those classes.
Despite this focus, some groups have trained unsupervised deep learning models for light
curve analysis, \citep{pasquet_pelican_2019, badura_recurrent_2022, kerr_using_2021, saha_rapid_2023}.
However, to our knowledge, our work here is the first to leverage unsupervised learned representations of light curves of SOs from a large dataset to downstream applications, i.e., anomaly detection, object characterisation, and synthetic data generation.
Figure \ref{fig:graphical_abstract} outlines the research in this study (into pre-training, fine-tuning, and generative AI capabilities). 

This paper is structured as follows: Section \ref{results} presents the results of our pre-training approach, anomaly detection, motion prediction fine-tuning, and synthetic data generation capabilities; Section \ref{sec:methods} details our methodological approach and implementation. Finally, Section \ref{outlook} discusses future directions and implications.

\section{Results \& Discussion}\label{results}
\subsection{Pre-training Perceiver-VAE to reconstruct MMT light curves}
\label{results:pre-training-training}
\begin{figure}[t]
    \centering
    \includegraphics[width=0.9\columnwidth]{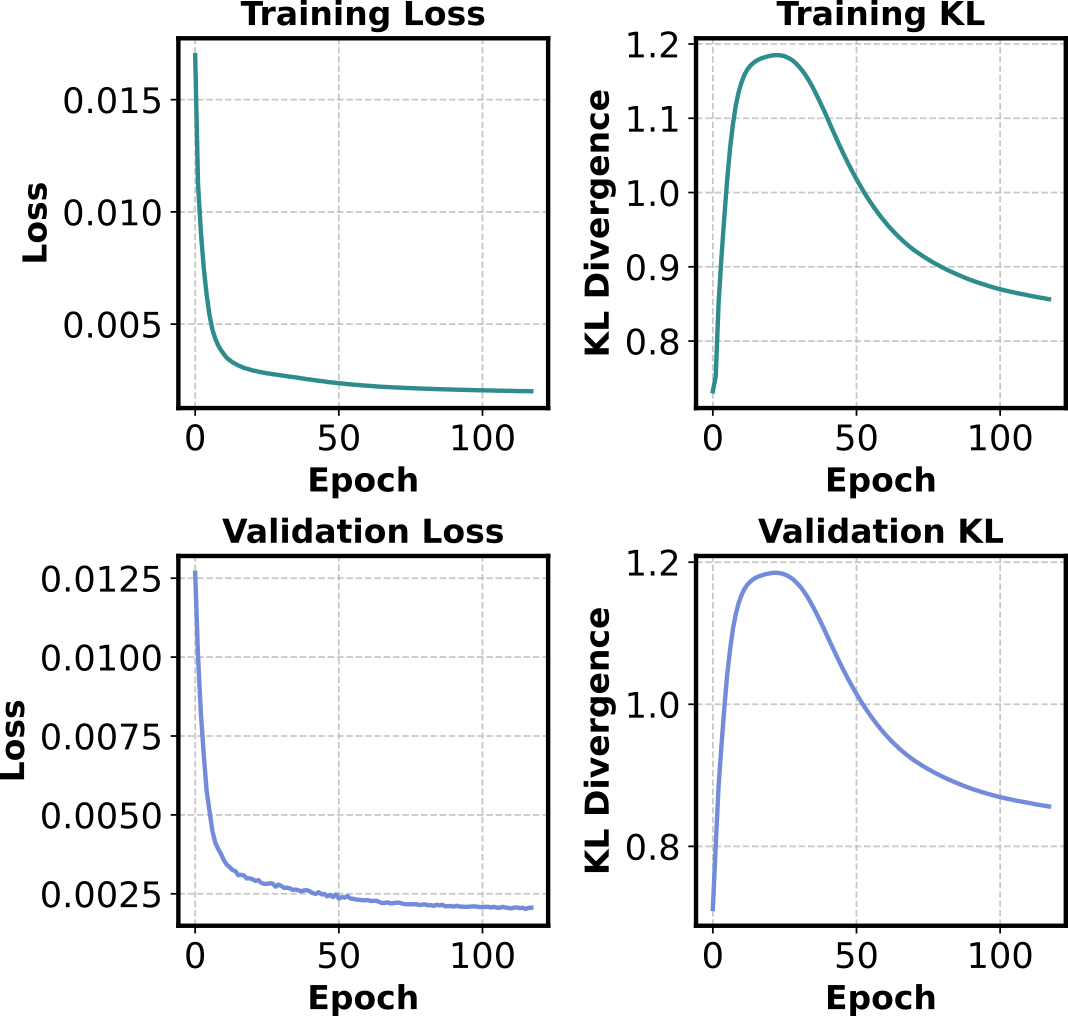}
    \caption{Training curves for a Perceiver-VAE model trained on the MMT-9 light curve dataset. Training and Validation losses decrease from approximately 0.015 to 0.0011 and 0.0012 respectively. The KL Divergence of the latent distribution initially increases, before plateauing at approximately 0.86. Validation loss decreases to a similar plateau, without the initial first epoch decrease.}
    \label{fig:train_loss_curves}
\end{figure}
First, we trained a VAE-based Perceiver (Perceiver-VAE) on the MMT-9 light curve dataset. Training showed expected convergence behaviour, with rapid initial improvement followed by gradual refinement. The reconstruction loss values reduce consistently before plateauing at 0.0011 (train) and 0.0012 (validation) (Figure \ref{fig:train_loss_curves}, Left-hand panels).
Additionally, when training a VAE, we optimise the latent space to be well ordered using a KL divergence term (see Equation \ref{eqn:kl_div}). This term measures how normally distributed the latent space is, and therefore can be interpreted as a measure of dataset heterogeneity.
The KL divergence terms exhibit interesting dynamics: they increase early in training, peaking at approximately epoch 25, then gradually reduce and stabilise.
This rise-then-fall dynamic occurs because we fixed the KL weight in our loss function as described in Methods Section \ref{methods:pre-training_strategy}, equation \ref{eqn:total_loss}.
This means early in training gradients from the self-supervised tasks (reconstruction, masking, forecasting) dominate, so the encoder expands the latent space, encoding maximal information. As reconstruction errors plateau, the KL term's relative contribution to the overall loss grows, so the fixed penalty increasingly pulls the latent distribution back towards the prior, causing the subsequent decrease.
The close tracking of train and validation loss suggest a good generalisation performance, without memorisation. Final KL Divergence values of approximately 0.86 indicate the model maintained a meaningfully ordered latent space whilst avoiding latent space collapse (recall that a KL divergence of 0 implies a normally distributed latent space, Figure \ref{fig:train_loss_curves}, right-hand panels). 
As a VAE loss function minimises the discrepancy between reconstruction and input data, the loss values can be interpreted directly as an error score. For our pre-training, the average validation error score was approximately 0.12\%.
\subsection{Flagging of anomalous Space Objects in the MMT-9 Dataset during pre-training}
\label{sec:pretraining_preflagging}
Using a held-out test portion of the MMT dataset, we examined the highest and lowest reconstruction error samples (Tables \ref{tab:opt_satellite_data_highest}-\ref{tab:satellite_data_lowest}). 
Among the highest reconstruction error test samples (Table \ref{tab:opt_satellite_data_highest}), we found a mixture of geodetic satellites, rocket bodies, debris, and Earth observation platforms. For example, light curves from EGS (AJISAI) appear 266 times in the MMT dataset, yet three light curves from this object appear among the ten highest reconstruction errors. Similarly, DELTA 1 DEB (180 appearances) and CZ-3B R/B (138 appearances) also showed high reconstruction errors despite relatively frequent representation in the pre-training data.
When an object such as AJISAI appears frequently in the pre-training 
data, the model learns a stable distribution of typical light curve 
morphologies for that object. A curve that falls outside this learned 
distribution is therefore more likely to represent atypical behaviour 
for that specific object, even if the object itself is well represented. 
For less frequently observed objects, the model has less prior 
information and cannot make the same distinction. This motivates the 
use of large, diverse pre-training datasets to maximise the range of 
objects for which such distinctions can be made.
For the lowest reconstruction error samples (Table~\ref{tab:satellite_data_lowest}), 
we again find a mixture of communications satellites, rocket bodies, Earth observation platforms, and military satellites, typically with fewer appearances in the pre-training data (ranging from 8-167).
\begin{center}
\begin{table*}[h]
\caption{The ten highest reconstruction errors from the pre-flagged light curve samples contained in the test set. We evaluated the lowest-loss model from Section \ref{results:pre-training-training} against the entire test dataset unseen during training/validation, recording the reconstruction error for each sample. The ten highest reconstruction error light curves represent a variety of space objects, including geodetic satellites, rocket bodies, debris, and Earth observation platforms. US: United States of America, JPN: Japan, PRC: People's Republic of China.\label{tab:opt_satellite_data_highest}}
\begin{tabular*}{\textwidth}{@{\extracolsep\fill}llllllll@{}}
\toprule
\textbf{Intl Code} & \textbf{NORAD} & \textbf{Name} & \textbf{Source} & \textbf{App.} & \textbf{Launch Date} & \textbf{Status} & \textbf{Notes} \\
\midrule
1986-061A & 16908 & EGS (AJISAI) & JPN & 266 & 1986-08-12 & In orbit & Geodetic \\
2019-078D & 44796 & CZ-3B R/B & PRC & 138 & 2019-11-23 & In orbit & Rocket Body \\
2015-003A & 40376 & SMAP & US & 50 & 2015-01-31 & In orbit & Earth Observation \\
1973-086U & 7028 & DELTA 1 DEB & US & 180 & 1973-11-06 & In orbit & Debris \\
1986-061A & 16908 & EGS (AJISAI) & JPN & 266 & 1986-08-12 & In orbit & Geodetic \\
1986-061A & 16908 & EGS (AJISAI) & JPN & 266 & 1986-08-12 & In orbit & Geodetic \\
2017-001B & 41912 & CZ-3B R/B & PRC & 45 & 2017-01-05 & Not in orbit & Rocket Body \\
1967-048A & 2807 & OPS 7218 (TRANSIT 16) & US & 93 & 1967-05-18 & In orbit & Navigation \\
2022-052AJ & 52630 & STARLINK-3989 & US & 8 & 2022-05-14 & In orbit & Communications \\
1972-058A & 6126 & LANDSAT 1 (ERTS 1) & US & 119 & 1972-07-23 & In orbit & Earth Observation \\
\bottomrule
\end{tabular*}
\end{table*}
\end{center}

\begin{center}
\begin{table*}[!h]
\caption{The ten lowest reconstruction error light curve samples from the test dataset. We evaluated the lowest-loss model from Section \ref{results:pre-training-training} against the entire test dataset unseen during training/validation, recording the reconstruction error for each sample. The ten lowest reconstruction error light curves similarly represent a variety of space objects, including Earth observation and communications satellites, rocket bodies, and military platforms. US: United States of America, JPN: Japan, PRC: People's Republic of China.\label{tab:satellite_data_lowest}}
\begin{tabular*}{\textwidth}{@{\extracolsep\fill}llllllll@{}}
\toprule
\textbf{Intl Code} & \textbf{NORAD} & \textbf{Name} & \textbf{Source} & \textbf{App.} & \textbf{Launch Date} & \textbf{Status} & \textbf{Notes} \\
\midrule
1999-051A & 25919 & IKONOS 2 & US & 46 & 1999-09-24 & In orbit & Earth Observation \\
2011-014A & 37386 & USA 229 & US & 167 & 2011-04-15 & In orbit & Satellite \\
2019-074AD & 44740 & STARLINK 1035 & US & 28 & 2019-11-11 & In orbit & Communications \\
2010-059A & 37214 & FENGYUN 3B & PRC & 133 & 2010-11-04 & In orbit & Weather \\
2003-054B & 28096 & ATLAS 2AS CENTAUR R/B & US & 167 & 2003-12-02 & In orbit & Rocket Body \\
2014-066B & 40287 & CZ-2C R/B & PRC & 39 & 2014-10-27 & In orbit & Rocket Body \\
2017-061D & 42958 & IRIDIUM 129 & US & 50 & 2017-10-09 & In orbit & Communications \\
2002-056A & 27597 & ADEOS 2 & JPN & 88 & 2002-12-14 & In orbit & Earth Resources \\
2022-053W & 52676 & STARLINK-4006 & US & 8 & 2022-05-18 & In orbit & Communications \\
1991-076C & 21799 & USA 74 & US & 93 & 1991-11-08 & In orbit & Satellite \\
\bottomrule
\end{tabular*}
\end{table*}
\end{center}

\subsection{Analysis of pre-training early-flagged Space Objects}
\label{sec:pre-flagged_anomalies_analysis}
\begin{figure*}[!h]
    \centering
    \includegraphics[width=0.8\textwidth]{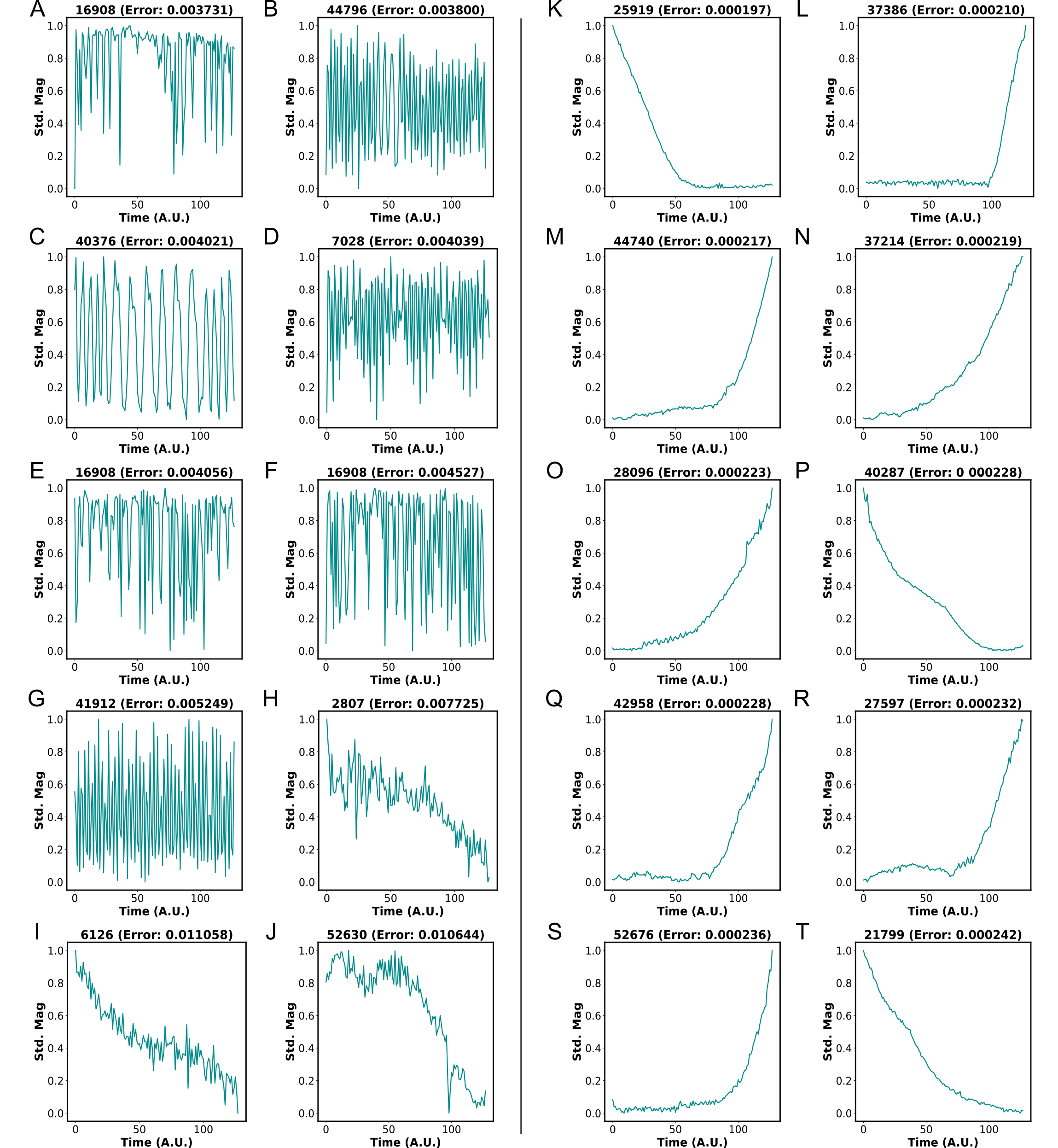}
    \caption
    {(A-J) The ten highest reconstruction error test light curves, where A is the highest, and J the tenth highest. These curves mostly exhibit periodic variation over time in the measured standardised magnitude. (K-T) As in A-J but for the ten lowest error test light curves, which exhibit mostly linear variation in the standardised magnitude over time.}
    \label{fig:ati:pre_flagged_anomalies}
\end{figure*}

To understand the nature of these highest/lowest reconstruction error curves, we manually inspected them (as shown in Figure \ref{fig:ati:pre_flagged_anomalies}). 
The ten highest reconstruction error curves were all periodic (Figure \ref{fig:ati:pre_flagged_anomalies} A-J), with frequent changes in magnitude. In contrast, the lowest error curves were substantially more linear (Figure \ref{fig:ati:pre_flagged_anomalies} K-T). Whilst most were monotonic, several curves (e.g., R, T) showed small deviations from this pattern. 

Highly periodic curves typically represent rapid tumbling behaviour of the space object as highlighted by \cite{qashoa_classification_2023}. 
Conversely, linear curves indicate objects with smaller attitude changes or highly regular morphology.
This analysis supports our presumption that tumbling behaviours in the highest reconstruction errors likely represent atypical SO behaviour.
Thus, frequent appearances in training data help the model flag what constitutes unusual behaviour for particular objects.

We note that reconstruction error may be partially confounded by observing conditions. The MMT-9 dataset does not provide per-point photometric noise or measures of uncertainty, so observing-condition noise and genuine morphological atypicality cannot be fully separated from reconstruction error alone as a ground-truth. Instead, we took mean observing distance as a proxy for signal-to-noise ratio. Analysing the full test set ($n = 24{,}995$), we found a small-to-moderate correlation between per-curve reconstruction error and mean observing distance, with Spearman $\rho = 0.34$ ($p < 0.001$), accounting for approximately 11\% of variance in reconstruction error (R$^2$ from Ordinary Least Squares on rank-transformed variables). 
This partial confound underscores the need for supervised fine-tuning (see Section~\ref{sec:anomaly_detection_fine-tuning}), where labelled examples allow the model to learn to distinguish genuine anomalies from observing-condition artefacts.
\subsection{Light curve forecasting}
\label{sec:forecasting}
An additional output from pre-training is forecasting future light curve states from past states.
By including forecasting as a self-supervised task (see Section \ref{methods:pre-training_strategy} for details), our model learns a latent space corresponding to different timecourse portions, enabling inference time predictions without  fine-tuning.
Analysing the same held out test dataset used in Section \ref{sec:pretraining_preflagging}, we computed the MSE forecasting loss after masking out the final 25\% of the timecourse (see Equation \ref{eqn:forecast_loss}). Across 24,995 test curves, the mean forecasting error was $7.7 \times 10^{-4}$, with a standard deviation of $7.6 \times 10^{-4}$.

\begin{figure}[!h]
    \centering
    \includegraphics[width=0.9\columnwidth]{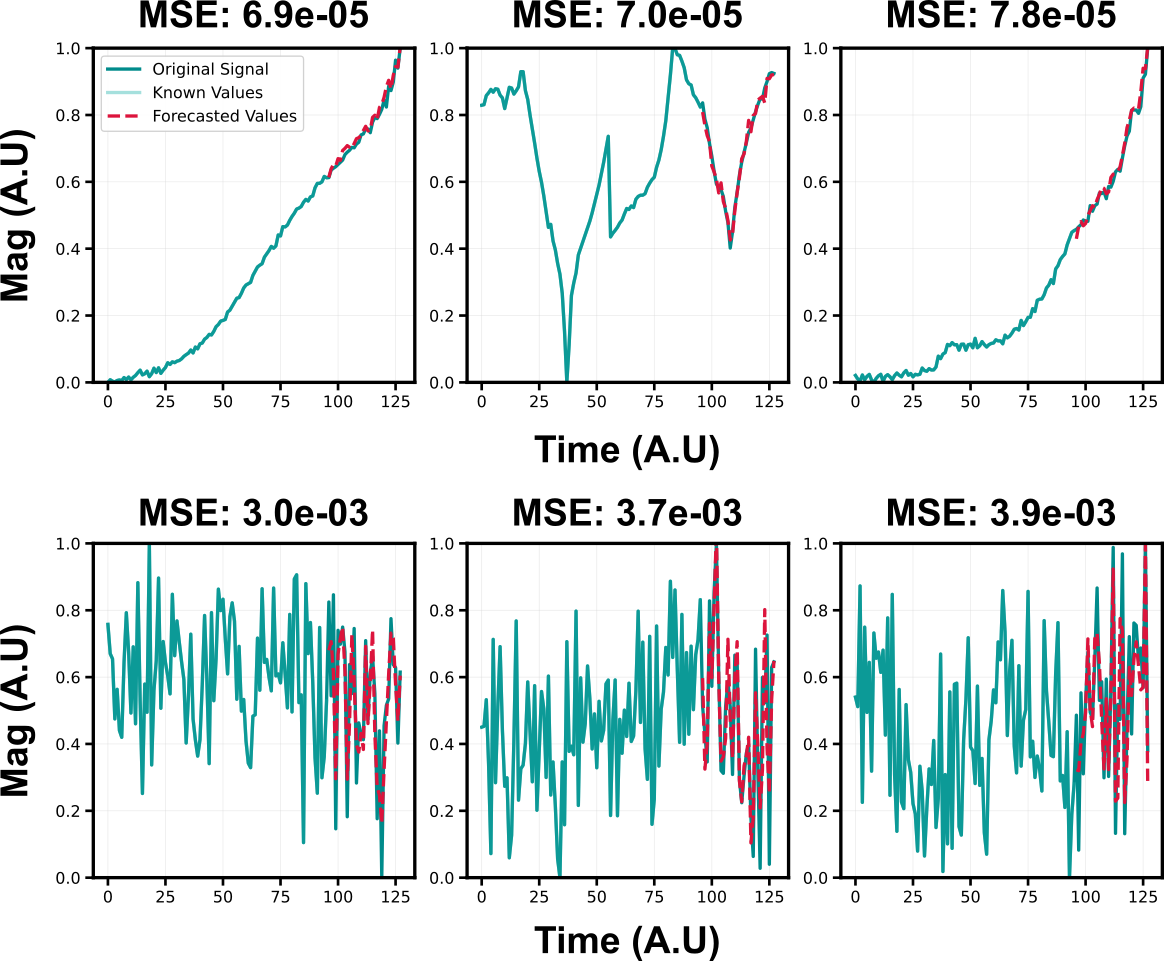}
    \caption{Test light curves with the latter 25\% of the timecourse masked at inference. Our Perceiver-VAE was trained with several self-supervised tasks, including a forecasting loss (see Section \ref{methods:pre-training_strategy} for details). This means that at inference, the future state of a light curve can be predicted by the pre-trained model. \textbf{Top row:} the lowest three forecasting mean squared error test samples. \textbf{Bottom row}: the highest three forecasting mean squared error test samples. Qualitatively, the original signal (turquoise) and known values (cyan) are aligned. In the lowest error samples, the forecasted values also align well with the original signal, capturing the dynamics well. In the highest error samples, the general trends of the masked regions are successfully forecasted, but the magnitude of the peaks/troughs in the signal is not fully captured.}
    \label{fig:forecasting}
\end{figure}

To examine the difference in a high/low forecasting error, we visualised the three lowest and highest errors (Figure \ref{fig:forecasting} top and bottom rows respectively).
In both cases, the general trends of the light curve were well forecasted (Figure \ref{fig:forecasting} red-dotted line). 
For the lowest error samples, our model accurately predicted future values with minimal deviation from ground truth.
For the highest error samples (predominantly high-frequency periodic curves), the model captures patterns well but struggles with exact magnitude prediction of these features (Figure \ref{fig:forecasting}).

This forecasting ability demonstrates that our model has learned meaningful temporal relationships in light curve dynamics unsupervised.
Across all the self-supervised learning (SSL) tasks, the reconstruction errors suggest our model has learned rich representations decodable to various light curve types and can potentially identify anomalies.
However, making definitive claims remains challenging with unlabelled data, as the exact events/behaviours causing normal/anomalous patterns are unclear.
Whilst step changes might represent manoeuvres or morphological features, validating these hypotheses requires supervised learning or physical modelling.
Therefore, we next examined fine-tuning with a synthetic, labelled dataset.
\subsection{Fine-tuning for anomaly detection}
\label{sec:anomaly_detection_fine-tuning}
\begin{figure*}[!t]
    \centering
    \includegraphics[width=0.8\textwidth]{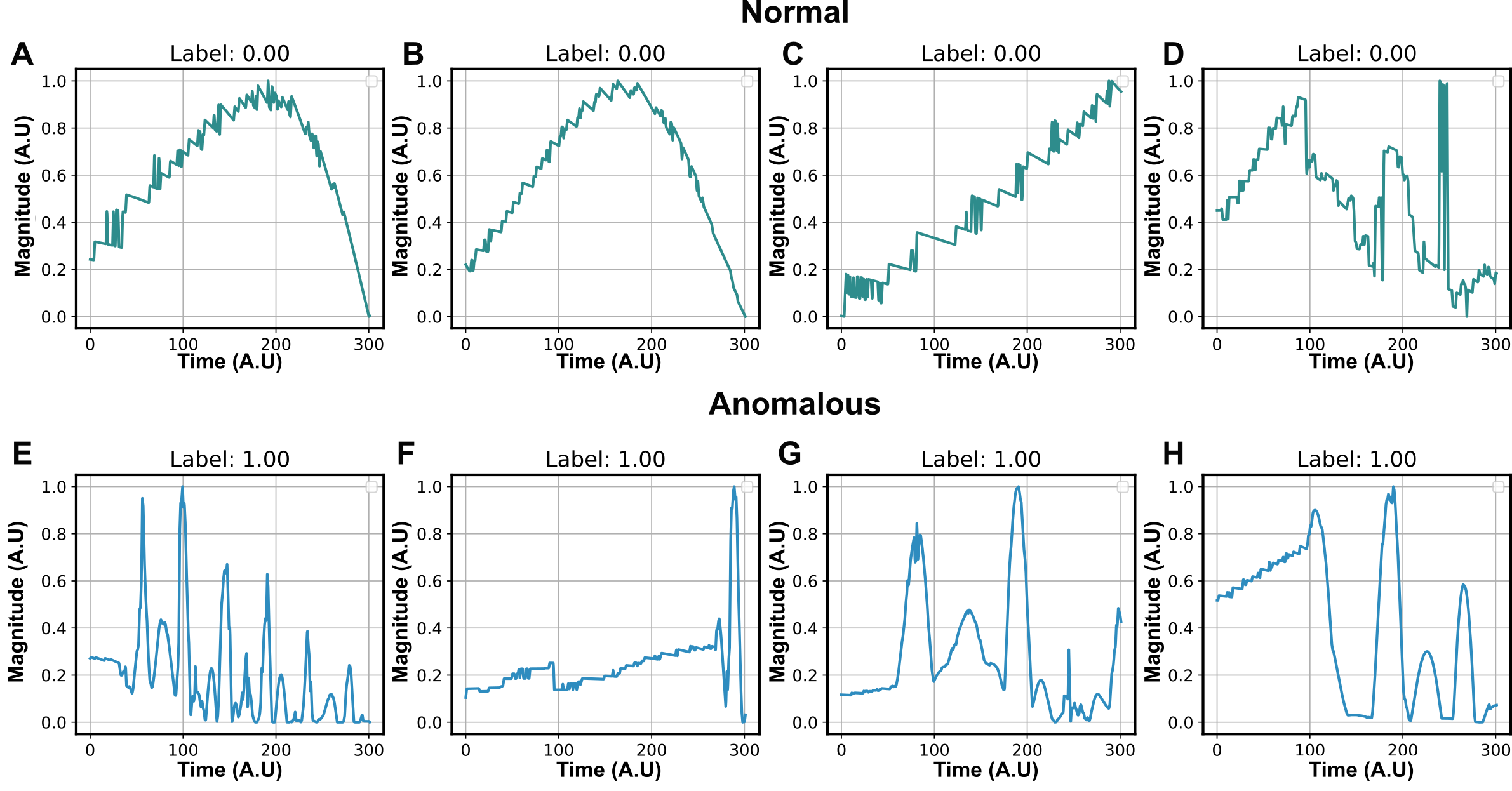}
    \caption
    {Example normal (A-D) and anomalous (E-H) light curves generated by CASSANDRA as a fine-tuning dataset. Normal curves are typically smooth bell-shaped profiles (A-B), monotonic increasing curves (C), or non-periodic variation with moderate changes in amplitude (D). Anomalous curves contain a variety of distinct characteristic features, such as high-frequency periodicity (E), isolated spikes on otherwise monotonic signals (F), irregular periodic profiles (G), and abrupt extended minima followed by a transition to a periodic profile (H).     }
    \label{fig:ati:fine-tuning_dataset}
\end{figure*}

Obtaining real observational data of satellites with known anomalies is extremely challenging, as the full set of possible anomalies is not well understood or catalogued, and organisations are opaque about anomalous space object behaviour for national security/commercial reasons.
To address this data gap, we produced a synthetic dataset for fine-tuning using the framework CASSANDRA (Computational Agent For Space Situational Awareness And Debris Remediation Automation). Within this framework is an orbital simulator first introduced by \cite{vasile_intelligent_2023}, generating the  light curves using a CAD model of an orbiting Space Object with a specific geometry.
This dataset contains 800 simulations of four different satellite platforms: a boxwing satellite (e.g., Jason-3), Sentinel-3, SMOS, and Starlink.
These space objects (SOs) orbit an observing ground station, with both light and spectral curves recorded as time-series. 
Within these 800 simulations, a subset contain anomalous events of varying magnitudes (specifically debris collisions that result in detectable changes in signal). These changes in signal are shown in Figure \ref{fig:ati:fine-tuning_dataset} F, H, where otherwise linear light curves abruptly change to a different profile, oscillating in magnitude between very high and low minima.
We first evaluated the anomaly detection capability of the largest  pre-trained model without 
any labelled data. To do this we passed a balanced set of 32 anomalous and 32 normal 
CASSANDRA light curves through the frozen encoder and used reconstruction error 
as an anomaly score, flagging the top 10\% highest errors as anomalous.
This unsupervised approach achieved 61\% accuracy and a ROC AUC of 0.71, demonstrating 
that the pre-trained representations capture anomaly-relevant structure without 
any supervision. As with our motion prediction fine-tuning dataset (Section~\ref{sec:fine-tuning_motion}), CASSANDRA is simulated, whereas pre-training uses real MMT-9 observations, introducing a domain gap between the two that we discuss further in Section~\ref{outlook}. Fine-tuning with labelled data helps bridge this domain gap, substantially improving performance, as described below.
For anomaly detection, we retain the frozen encoder throughout; end-to-end fine-tuning was evaluated (i.e., where the encoder's weights were updated in fine-tuning), but did not improve performance for the anomaly detection task.
To fine-tune our pre-trained Perceiver-VAE for anomaly detection, we 
freeze the encoder weights and train a lightweight MLP classification 
head on the learned latent representations (see 
Section~\ref{sec:fine-tuning_methods} for details).

We compare against four supervised baselines trained from scratch on 
raw light curves, as well as randomly initialised Perceiver encoders 
(r.e.) at increasing scale (Table~\ref{tab:anomaly_comparison}). Among 
the baselines, the CNN achieves the strongest ROC AUC (0.849), with the 
LSTM achieving the best accuracy (0.856). Randomly initialised 
Perceivers are broadly competitive with these baselines, suggesting 
the architecture itself has useful inductive biases for this task. 
Pre-training consistently improves on random initialisation across all 
scales, with pre-trained models achieving ROC AUC of 0.911--0.916. Our 
full model achieves 0.850 accuracy, 0.916 ROC AUC, and 0.793 macro F1, 
outperforming all baselines across metrics.

To assess label efficiency, we evaluated the pre-trained Perceiver 
against the CNN baseline in a $k$-shot regime, training each model on 
$k \in \{2, 5, 10, 25\}$ labelled examples per class across 20 
independent runs (Table~\ref{tab:anomaly_data_efficiency}). The 
pre-trained Perceiver outperforms the CNN at all values of $k$, 
including as few as two labelled examples per class (ROC AUC 
$0.777 \pm 0.082$ vs.\ $0.731 \pm 0.049$). The advantage of 
pre-training grows with $k$, reaching a gap of $0.082$ AUC points at 
$k=25$. This demonstrates that the representations learned during 
pre-training are particularly valuable in the label-scarce regime 
typical of anomaly detection in operational SDA settings.

\begin{figure*}[!t]
    \centering
    \includegraphics[width=0.7\textwidth]{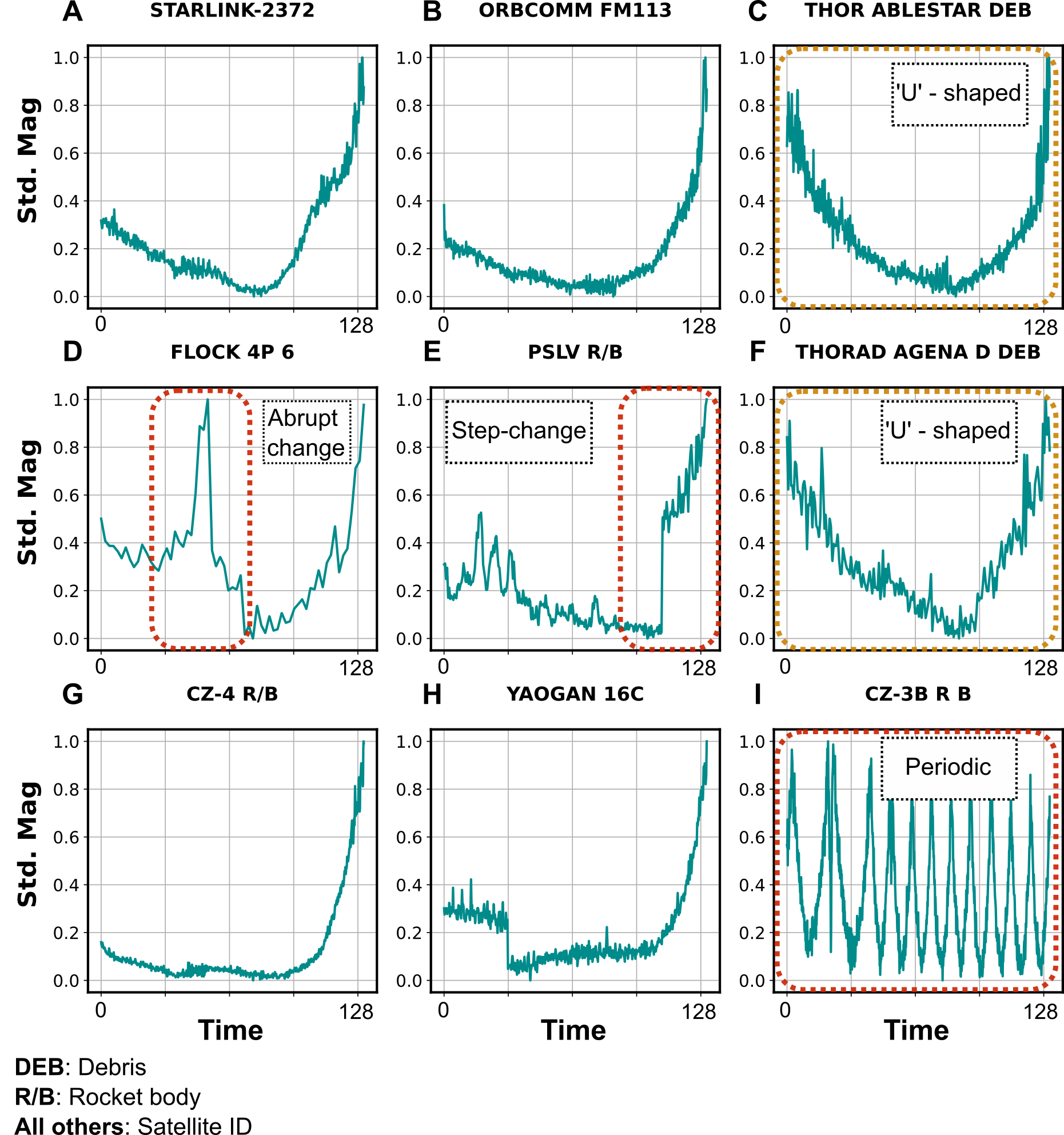}
    \caption
    {Example high-confidence anomaly predictions (probability $\geq 0.95$) from the fine-tuned anomaly detection light curve Perceiver-VAE model on an independent test set of MMT light curves. We mapped the light curves back to the names of the object, finding a mixture of satellites (A, B, D, H), debris (C, F), and rocket bodies (E, G, I). Within these predicted anomalies, we see notable anomalous patterns such as abrupt peak changes in magnitude (D, red dotted outline), step changes (E, red dotted outline), and highly periodic curves (I, red dotted outline). In addition to well documented anomaly profiles, we also identified characteristic debris profiles. Here  debris light curves have a symmetric 'U' shaped profile, where the minima in magnitude lies approximately in the midpoint of the observation (C, F, orange dotted outlines). Units: Std. Mag: Standardised Magnitude rescaled between 0 and 1 (A.U.). Time: Resampled to 128 observations (A.U.).
    }
    \label{fig:ati:fine-tuning_predictions}
\end{figure*} 

\begin{table}[h!]
\caption{Anomaly detection performance on the held-out test set. Supervised baselines are trained from scratch on raw light curves. Perceiver models are shown at increasing scale; random encoder (r.e.) variants confirm that pre-trained representations drive performance.\label{tab:anomaly_comparison}}
\centering
\resizebox{\columnwidth}{!}{%
\begin{tabular}{llccc}
\toprule
\textbf{Model} & \textbf{Size} & \textbf{Accuracy} & \textbf{ROC AUC} & \textbf{Macro F1} \\
\midrule
RNN            & -- & 0.831 & 0.635 & 0.587 \\
LSTM           & -- & 0.856 & 0.741 & 0.783 \\
CNN            & -- & 0.819 & 0.849 & 0.743 \\
Transformer    & -- & 0.800 & 0.774 & 0.672 \\
\midrule
Perceiver (r.e.)  & s & 0.756 & 0.747 & 0.559 \\
Perceiver (r.e.)  & m & 0.756 & 0.734 & 0.559 \\
\midrule
Perceiver (pre., frozen) & s & 0.825 & 0.911 & 0.763 \\
Perceiver (pre., frozen) & m & 0.856 & 0.912 & 0.796 \\
\midrule
\textbf{Ours (pre., frozen)} & \textbf{l} & \textbf{0.850} & \textbf{0.916} & \textbf{0.793} \\
\bottomrule
\end{tabular}}
\end{table}

\begin{table}[h!]
\caption{Label efficiency comparison for anomaly detection. ROC AUC 
(mean $\pm$ std. dev. across 20 independent runs) at increasing numbers of 
labelled training examples ($k$ per class). The pre-trained Perceiver 
(large) outperforms the CNN baseline at all values of $k$.\label{tab:anomaly_data_efficiency}}
\centering
\resizebox{\columnwidth}{!}{%
\begin{tabular}{lcccc}
\toprule
\textbf{Model} & \multicolumn{4}{c}{\textbf{ROC AUC by }$k$\textbf{ shots per class}} \\
\cmidrule(lr){2-5}
 & $k=2$ & $k=5$ & $k=10$ & $k=25$ \\
\midrule
CNN (baseline)   & 0.731 $\pm$ 0.049 & 0.754 $\pm$ 0.034 & 0.754 $\pm$ 0.009 & 0.780 $\pm$ 0.008 \\
Perceiver (pre.) & 0.777 $\pm$ 0.082 & 0.800 $\pm$ 0.109 & 0.851 $\pm$ 0.040 & 0.862 $\pm$ 0.030 \\
\bottomrule
\end{tabular}}
\end{table}

Having fine-tuned our model to high classification accuracy, we next analysed the held-out test dataset, classifying test samples as anomalous/non-anomalous and visualising example confidently predicted anomalies setting a probability threshold of $\geq 0.95$ (Figure \ref{fig:ati:fine-tuning_predictions}).
We found a variety of SOs in this set, including several satellites (Figure \ref{fig:ati:fine-tuning_predictions} A, B, D, H). All of these satellites except the light curve belonging to FLOCK 4P 6 (Figure \ref{fig:ati:fine-tuning_predictions} D) exhibited 'J' shaped curves, which simulations suggest may represent glinting from a highly reflective surface aligning with the sun \citep{matsushita_light_2019}. 
We also detected light curves belonging to debris (Figure \ref{fig:ati:fine-tuning_predictions} C, F), which displayed prominent 'U' shaped profiles i.e., where the minima in magnitude is approximately in the middle of the observation. Finally, we identified three other distinct light curve motifs (Figure \ref{fig:ati:fine-tuning_predictions} red dotted outlines): abrupt changes in magnitude which presented as peaks (Figure \ref{fig:ati:fine-tuning_predictions} D), step changes (Figure \ref{fig:ati:fine-tuning_predictions} E), or highly periodic variation (Figure \ref{fig:ati:fine-tuning_predictions} I).  
We note that whilst these predictions are made on real MMT-9 observations, no ground truth anomaly labels exist for this dataset and therefore real anomalies cannot be confirmed this way. The identified morphologies are therefore interpreted qualitatively, guided by known physical signatures from the literature and simulation. Systematic validation against catalogued anomaly events or operator records remains an important direction for future work.
\subsection{Fine-tuning for motion prediction}
\label{sec:fine-tuning_motion}
\begin{figure}[!t]
    \centering
    \includegraphics[width=0.8\columnwidth]{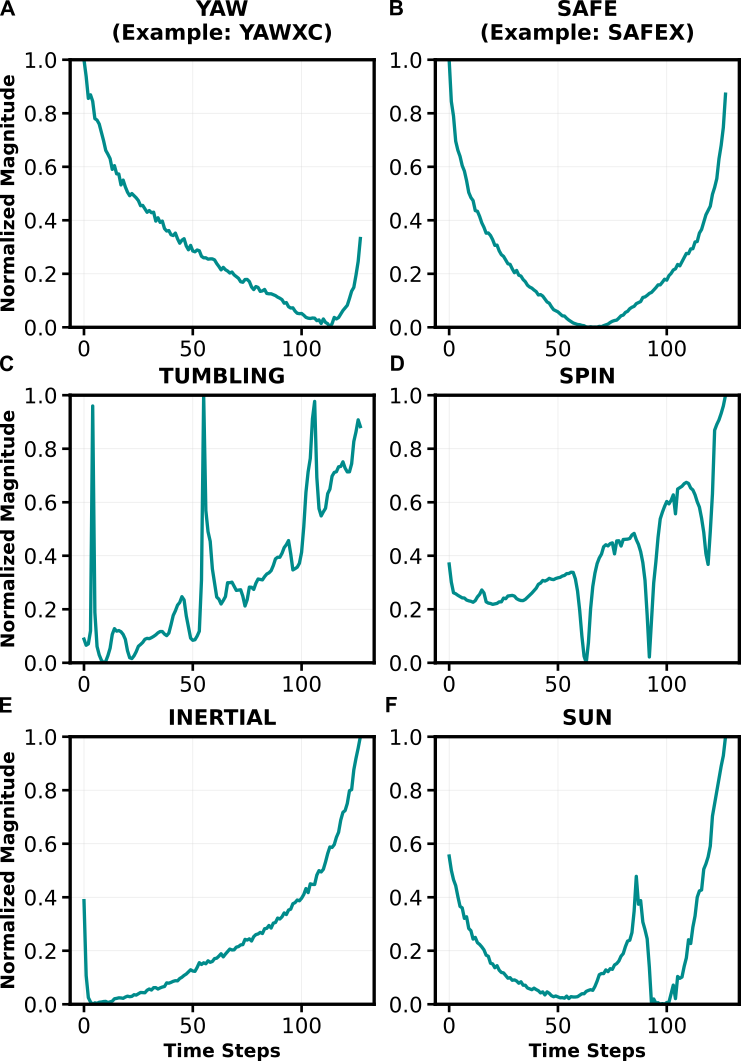}
    \caption
    {Example light curves from the GMV GRIAL motion fine-tuning dataset. (A-F) Six grouped motions are labelled into classes. (A) YAW (grouped) YAWXC, YAWXS. The Yaw motions represent the satellite pointing its X axis to nadir, compensating for either optical distortion (XC), or maximising solar array lighting (XS). (B) SAFE (grouped) SAFEX, SAFEZ. The Safe motions represent celestial North pole pointing, with either its X (SAFEX) or Z axis (SAFEZ) pointing to the Sun. (C) Tumbling, representing uncontrolled/complex rotation. (D) SPIN, representing simpler axial rotation. (E) INERTIAL, fixed pointing, and (F) SUN, representing the space object's X axis pointing to the sun, with Y to the celestial North pole, differentiated from SAFEZ by an X-axis phase angle.}
    \label{fig:ati:fine-tuning_motion_examples}
\end{figure}

We also fine-tuned on a separate dataset with ground-truth motions, simulated by the GRIAL simulator from GMV \citep{gallego_rso_2023}. Example light curves from this dataset are visualised in Figure \ref{fig:ati:fine-tuning_motion_examples}, here we briefly summarise the dataset.

\begin{table}[h!]
\caption{Comparison of motion classification performance on the held-out test set. Supervised baselines are trained from scratch on raw light curves. Perceiver models are shown at increasing scale; random encoder (r.e.) variants use randomly initialised weights to confirm that pre-trained representations drive performance. 'frozen' indicates the pre-trained encoder is not updated during fine-tuning (only the classification head is trained), whilst 'e2e' indicates the encoder is unfrozen and updated jointly with the head; Our full model (end-to-end fine-tuning) achieves best performance.\label{tab:baseline_comparison}}
\centering
\resizebox{\columnwidth}{!}{%
\begin{tabular}{llccc}
\toprule
\textbf{Model} & \textbf{Size} & \textbf{Accuracy} & \textbf{ROC AUC} & \textbf{Macro F1} \\
\midrule
RNN            & -- & 0.354 & 0.686 & 0.273 \\
Transformer    & -- & 0.521 & 0.753 & 0.355 \\
LSTM           & -- & 0.693 & 0.920 & 0.652 \\
CNN            & -- & 0.711 & 0.911 & 0.644 \\
\midrule
Perceiver (r.e.)  & s & 0.471 & 0.625 & 0.199 \\
Perceiver (r.e.)  & m & 0.432 & 0.500 & 0.101 \\
\midrule
Perceiver (pre., frozen) & s & 0.635 & 0.860 & 0.541 \\
Perceiver (pre., frozen) & m & 0.638 & 0.871 & 0.556 \\
Perceiver (pre., frozen) & l & 0.733 & 0.907 & 0.628 \\
\midrule
\textbf{Ours (pre., e2e)} & \textbf{l} & \textbf{0.827} & \textbf{0.951} & \textbf{0.773} \\
\bottomrule
\end{tabular}}
\end{table}

The GMV dataset is a simulated dataset of 22,006 light curves of Sentinel-3A simulated under 10 different motion laws as seen from 30 different ground stations.
These motions can be grouped into three distinct behaviours: First, Sun oriented: \textbf{SafeX}, \textbf{SafeZ} (where the Space Object's X/Z axis points to the Sun respectively, and Y initially points to the celestial North pole, and the X/Z  rotates respectively), and \textbf{Sun} (X-axis points to Sun, Y-axis to Celestial North pole, differentiated from Safe by a varied X-axis phase angle). Next, Earth oriented motions: \textbf{YAWXC, YAWZC, YAWXS, YAWZS} (where X/Z indicates the Space Object's nadir pointing axis, and C/S indicates whether the object's motion is compensating (C) for optical distortion, or (S) maximising solar array lighting).
Finally, the dataset also contains labelled general motions less related to satellite function: Tumbling (uncontrolled/complex rotation of the SO), Spin (rotation around a single axis), Inertial pointing (where the SO has a fixed orientation relative to the J2000 reference frame).
To fine-tune our pre-trained Perceiver-VAE, we first group similar motions together, as domain expert manual inspection between e.g., different yaw axial motions is a substantially more challenging task than distinguishing between different motions altogether (e.g., tumbling/sun pointing). 

Table \ref{tab:baseline_comparison} compares our model against supervised baselines and randomly initialised Perceiver-VAEs of varying sizes. 
Among the baselines, the LSTM and CNN models achieve the strongest performance (ROC AUC 0.92 and 0.91 respectively), with both the RNN (ROC AUC 0.69) and Transformer (ROC AUC 0.75) weaker on this task. Similarly, randomly initialised perceiver encoders perform poorly (ROC AUC 0.63 and 0.50 for the small and medium variant respectively). Pre-training provides substantial improvements for feature extraction at all scales, with the pre-trained Perceiver-VAEs reaching 0.86-0.91 ROC AUC. Our full large model uses end-to-end fine-tuning of the pre-trained encoder alongside the classification head (see Section \ref{sec:fine-tuning_methods} for details), which further increases classification metrics (0.95 ROC AUC, with 83\% accuracy and 0.77 F1 score).

This end-to-end trained Perceiver-VAE (Table~\ref{tab:baseline_comparison}, pre., e2e) was trained for up to 150 epochs (with early stopping, see Section \ref{sec:fine-tuning_methods} for full details); full per-class performance is shown in Table~\ref{tab:motion_classification_metrics}.
There was inter-class variation in the fine-tuned classifier's performance. For example, both tumbling and spin show high performance, with precision of approximately 0.97 and 0.90 respectively. Similarly, these classes also showed high recall, both with approximately 0.93. In contrast, inertial has high precision (0.86) but low recall (0.33). This is potentially due to morphological similarity between inertial and yaw light curves (the largest class), both of which exhibit smooth, monotonically increasing profiles without substantial local variation in magnitude (Figure \ref{fig:ati:fine-tuning_motion_examples}, compare A and E).

Further, INERTIAL misclassifications are concentrated specifically in the YAW class rather than distributed across all classes (Supplementary Figure~\ref{fig:supp:confusion_matrix}), consistent with feature overlap. Other under-represented classes, e.g., SUN (the smallest class $n_{\text{train}}=1{,}253$), achieve a substantially higher recall (0.632) than INERTIAL ($n_{\text{train}}=1{,}679$, recall 0.328), despite having fewer training examples. This supports the hypothesis that morphological overlap with YAW, rather than the number of INERTIAL training examples, is the primary driver of the low recall.
\begin{center}
\begin{table*}[h!]
\caption{Performance evaluation of the motion classification model on a held-out test dataset. The model achieves 82.74\% overall accuracy with an average ROC AUC of 0.9512. The metrics demonstrate varied performance across different motion classes, with TUMBLING and SPIN showing the strongest classification performance, whilst INERTIAL exhibits high precision but low recall. For YAW, the most common class, the model shows high recall but lower precision. $n_{\text{train}}$ denotes the number of training samples per class. \label{tab:motion_classification_metrics}}
\begin{tabular*}{\textwidth}{@{\extracolsep\fill}lcccccr@{}}
\toprule
\textbf{Motion Class} & \textbf{Precision} & \textbf{Recall} & \textbf{F1-Score} & \textbf{ROC AUC} & \textbf{\textit{n\textsubscript{test}}} & \textbf{\textit{n\textsubscript{train}}} \\
\midrule
INERTIAL & 0.860 & 0.328 & 0.475 & 0.869 & 262 & 1,679 \\
SAFE     & 0.771 & 0.749 & 0.760 & 0.948 & 342 & 2,586 \\
SPIN     & 0.900 & 0.929 & 0.915 & 0.991 & 225 & 1,707 \\
SUN      & 0.676 & 0.632 & 0.653 & 0.945 & 152 & 1,253 \\
TUMBLING & 0.970 & 0.928 & 0.949 & 0.997 & 278 & 1,778 \\
YAW      & 0.811 & 0.972 & 0.885 & 0.957 & 942 & 6,841 \\
\midrule
\textbf{Overall} & -- & -- & -- & \textbf{0.951} & \textbf{2201} & \textbf{15,844} \\
\textbf{Accuracy} & \multicolumn{3}{c}{\textbf{0.827}} & & & \\
\bottomrule
\end{tabular*}
\end{table*}
\end{center}
\subsection{Synthetic data generation}
\label{sec:synthetic_data_generation}
Alongside anomaly detection and motion prediction, we can test the utility of our learned embeddings in producing synthetic data.

The generation of useful and representative training datasets for Space Object (SO) behavioural analysis presents us with several challenges. First, real-world observations are limited by various constraints, such as telescope location, atmospheric conditions etc.
Because of this, collecting datasets which are comprehensive across all the various operational modes, manoeuvres and possible anomalous behaviours requires an unrealistic amount of observation time and resource use.
Simulators can partially address these challenges but are often constrained by computational costs and the reliability of the underlying physical model.

For these reasons, a method of generating a large amount of diverse physically possible synthetic light curves would substantially enhance training datasets for downstream SDA tasks.
Whilst several numerical light curve simulators exist (e.g., we use two distinct software to produce fine-tuning light curves in this study), producing light curves numerically often requires a large amount of priors and a long time to simulate. For example CASSANDRA produces approximately 1000 light curves per hour of simulation. In contrast, producing novel synthetic light curves from the Perceiver-VAE model only requires decoding the learned latent vectors, which is computationally undemanding. This process scales easily, and our initial unoptimised tests presented below generate  approximately 40,000 light curves per hour.

Recall that training a Variational Autoencoder encodes a continuous latent space, from which we can sample and decode to generate new data points.
In addition to enforcing a well-structured latent space, this continuity also implies a functionally infinite number of decodable latent vectors, and therefore also of synthetic data generation.

%
There are two key considerations when selecting the latent vectors to generate synthetic data. First, physical possibility/plausibility, which ensures generated curves represent possible SO behaviours rather than artefacts of the autoencoding process. Second, the utility of the synthetic data, whereby we want to generate synthetic data that is useful to the field of SDA.
To address these concerns, we outline a methodical approach below which emphasises constraining the generation process over sampling in a more general or random way. In other words, potentially the most simple method would be to sample randomly from the learned embedding space, and decode those into light curves. However, whilst this can produce light curves that look roughly realistic, there is no way to validate whether these light curves are physically possible, or to interpret what they represent. 

This is a particularly salient concern, as our Perceiver-VAE is pre-trained on many different Space Objects (SOs), including multiple types (e.g., satellites, rocket bodies, debris).
It is conceivable that we could index a set of latent vectors which are halfway between encoding a Sentinel-3A light curve and a piece of debris, which is an impossible curve.

\begin{figure*}[!h]
    \centering
    \includegraphics[width=0.9\textwidth]{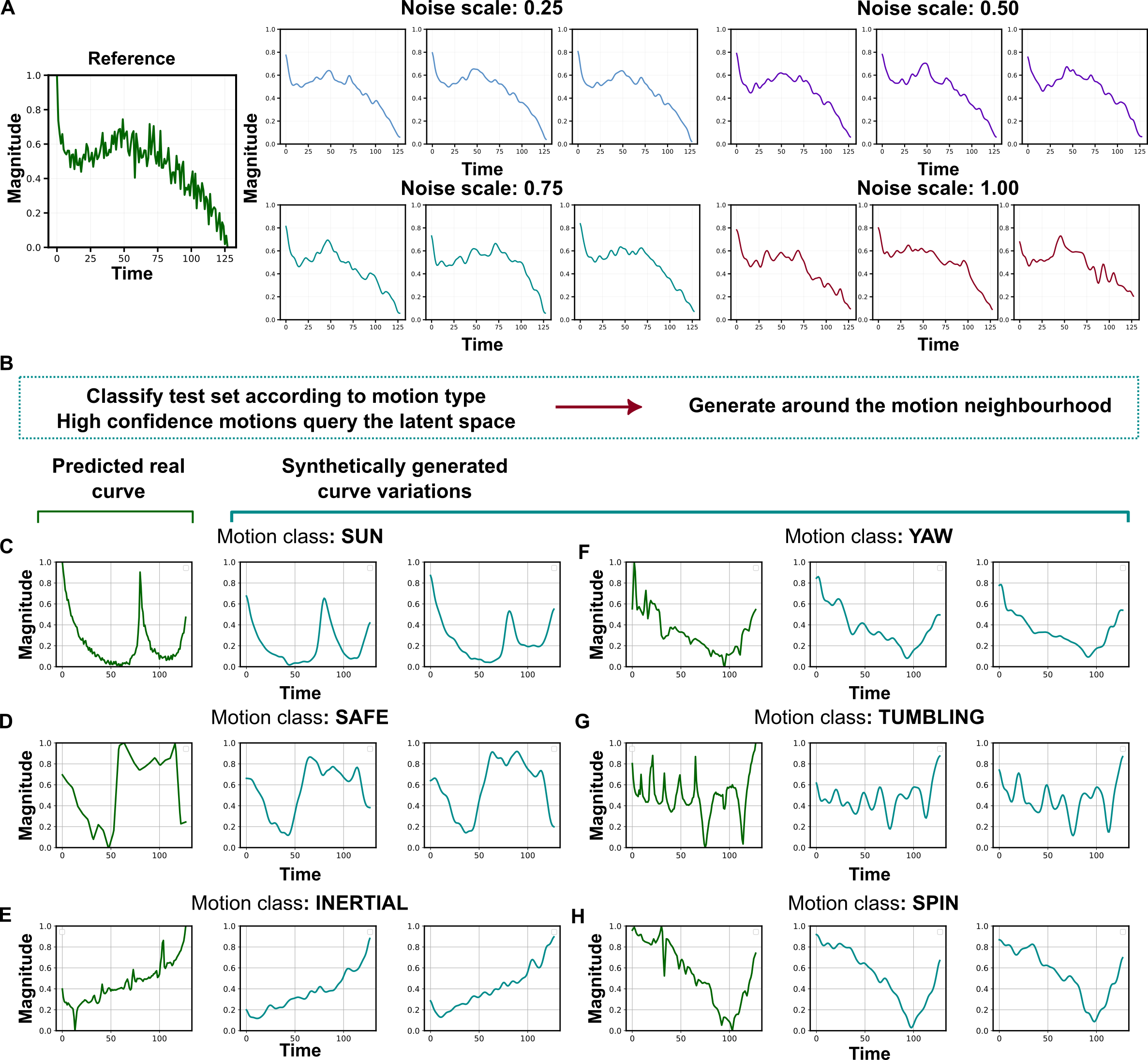}
    \caption
    {Sampling the latent space of our fine-tuned Perceiver-VAE. \textbf{(A)} Initial exploration of noise scaling in synthetic data generation. As in VAE training, at inference time we can offer a reference real light curve into the forward pass of the model. Once this has been encoded, we  capture the latent vectors that respond highest to this reference curve, apply Gaussian noise at different scales to these scalar values (0.25-1.0), and then decode the results. As expected, this results in synthetic light curves that adhere to the general global feature of the input curve, but differ on local features, and at increasing variance as you increase the distance from the latent activation. \textbf{(B)} Our synthetic data generation querying approach. Here we use the fine-tuned motion classifier, but this could also readily be the anomaly tuned classifier from Section \ref{sec:anomaly_detection_fine-tuning}. \textbf{(C-H)} To generate informed synthetic data, we first classify a held out test set using the motions described in Figure \ref{fig:ati:fine-tuning_motion_examples}. Selecting a desired motion (e.g., Tumbling), we isolated test samples with high confidence class likelihood of belonging to tumbling, before using those as our latent space activating samples. As in (A), we then generate synthetic samples using Gaussian noise with noise scale of 0.75 around the latent 
activations.}
    \label{fig:ati:syndata}
\end{figure*}

Instead, we developed a reference-based sampling approach that ties the synthetic data generation to known physical behaviours whilst allowing for controlled variation. To do this, we queried our embedding space using reference curves as inputs (Figure \ref{fig:ati:syndata}). First, we provided an arbitrary index (Figure \ref{fig:ati:syndata}A), sampling the latent vectors which most strongly activated in response to that input curve, adding varying amounts of noise to those activation vectors, and decoding the resulting vectors into light curves.

Figure \ref{fig:ati:syndata} shows that we can produce curves with a similar overall trend to the reference curve (i.e., that are appropriately constrained by the reference curve).
Secondly, we show that adding increasing amounts of noise to these activations produces curves which vary further from the reference curve (compare Figure \ref{fig:ati:syndata}A, noise scale 1.00 curves to Figure \ref{fig:ati:syndata}A, noise scale 0.25 curves).

Next, we used the fine-tuned model from Section \ref{sec:fine-tuning_motion} to generate motion mode informed synthetic light curves (Figure \ref{fig:ati:syndata}B). 
Figure~\ref{fig:ati:syndata}C-H shows example synthetic curves generated for each of the six motion classes, alongside the real reference curve used to seed generation for that class.
To do this, we first took the held-out test set which we used to evaluate the fine-tuned model from Section \ref{sec:fine-tuning_motion}. Next we obtained reference curves for the latent space by using light curves with high confidence predictions from that model (>95\% likelihood). 
Finally, we generated around the neighbourhood of those references by first applying noise to those latent activations and then decoding into light curves as in Figure \ref{fig:ati:syndata}A. 

\begin{figure*}[htbp]
    \centering
    \includegraphics[width=0.8\textwidth]{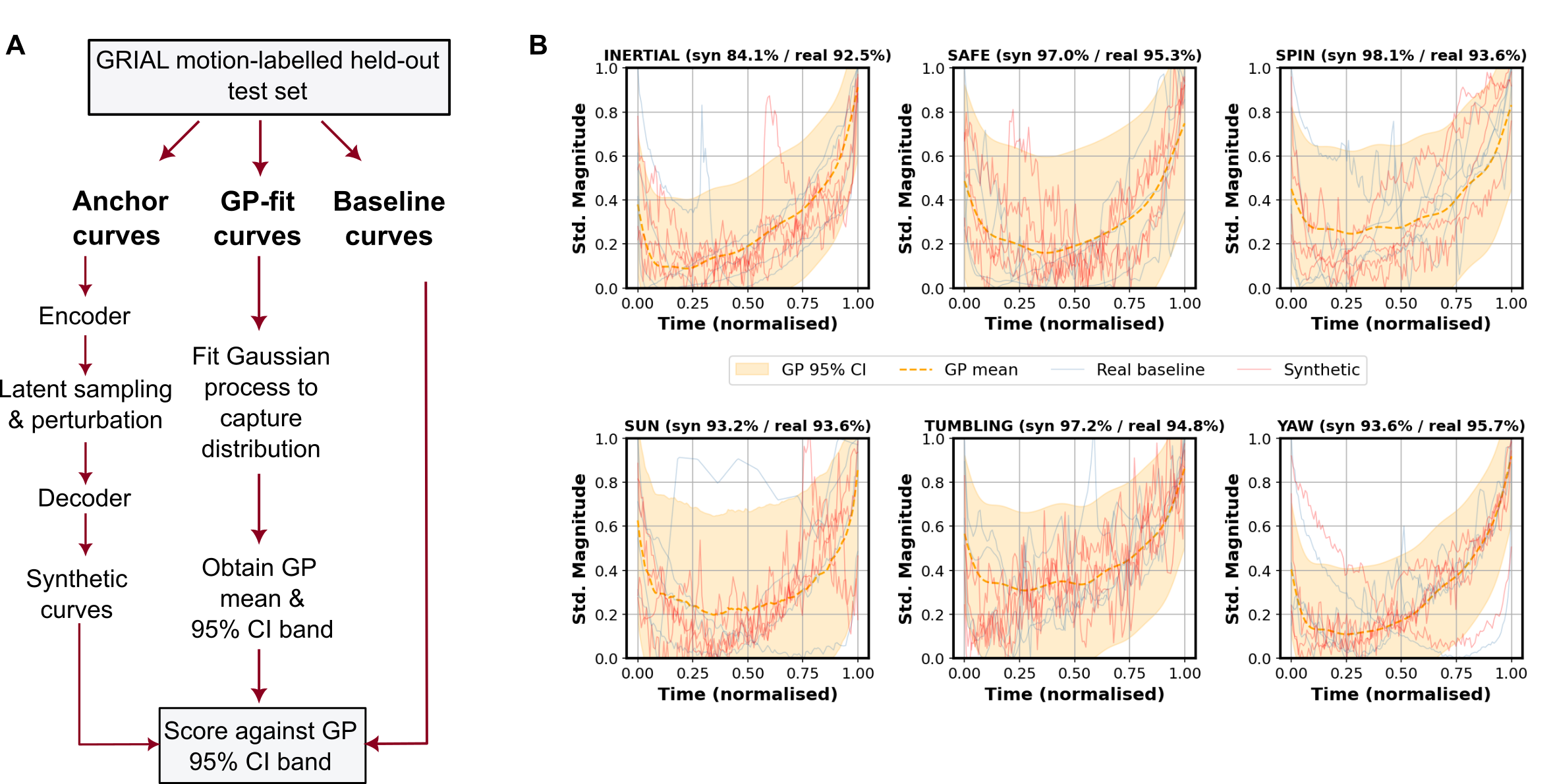}
\caption{Gaussian Process validation of synthetic light curves. 
\textbf{(A)} Overview of the validation procedure. From the held-out test set, three subsets are selected per class: anchor curves (used to seed synthetic generation), GP-fit curves (used to fit the GP), and baseline curves (scored against the fitted GP as a reference for natural sampling variation). Synthetic curves are generated using the Perceiver-VAE encoder-decoder as described in Section~\ref{sec:syndata_methods} and scored against the resulting confidence band alongside the real baseline curves. \textbf{(B)} GP validation panels for all six motion classes. The orange band and dashed line show the GP 95\% confidence interval and mean respectively. Real baseline curves are shown in blue and synthetic curves in red. Each panel title reports the fraction of time steps within the GP 95\% CI, averaged across synthetic curves (syn) and real baseline curves (real) respectively. Synthetic scores range from 84.1\% (INERTIAL) to 98.1\% (SPIN), comparable to real baseline scores (92.5--95.7\%), with the gap small and bidirectional across classes, consistent with sampling variation rather than a systematic generation artefact.}
    \label{fig:gp_validation}
\end{figure*}

Without real labelled light curves to compare against, validating these synthetic data is challenging. To this end, we provide a quantitative assessment of synthetic light curve quality via validation against a Gaussian Process (GP) fitted to the real light curve distribution for each motion class (Figure~\ref{fig:gp_validation}). The held-out test split was partitioned into three subsets per class using true class labels (see Section~\ref{sec:syndata_methods} for the full partitioning 
procedure). Synthetic and baseline curves were  scored by the fraction of time steps falling within the GP's 95\% confidence interval.
Synthetic scores ranged from 84.1\% (INERTIAL) to 98.1\% (SPIN), comparable to real baseline scores (92.5--95.7\%) across all six classes, with the gap between synthetic and real scores small and bidirectional across classes, consistent with sampling variation rather than a systematic generation artefact. These results demonstrate that the synthetic curves produced by our neighbourhood sampling approach are statistically consistent with real light curve distributions across motion classes.

From this, we saw that in each motion class, we could in essence query an inference dataset to produce more curves of a similar attitude motion mode. Considering that producing these fine-tuning data is time and computationally expensive, we present this methodology as a potentially valuable tool for the SDA field.

\subsection{Model Compression via Magnitude Pruning and Quantisation}
\label{sec:pruning}
To assess parameter efficiency of the large Perceiver-VAE, we applied magnitude 
pruning and INT8 quantisation to the motion classification fine-tuned model, 
targeting the Perceiver layers and the VAE bottleneck separately for pruning. Beyond BatchNorm recalibration (Section~\ref{sec:compression_methods}), no 
further fine-tuning was performed after magnitude pruning. 
For quantisation, PyTorch's dynamic quantisation computes quantisation parameters at inference time and requires no calibration data.
The Perceiver 
layers tolerate up to 70\% weight sparsity with negligible degradation in motion 
classification accuracy (82.5\% vs.\ 83.0\% baseline). In contrast, the VAE 
bottleneck is considerably more sensitive: 50\% sparsity reduces accuracy to 
77.2\%, a drop of approximately 6 percentage points. This asymmetry suggests 
that whilst the Perceiver layers contain substantial redundancy at this scale, 
the bottleneck representations are dense and critical to downstream performance. 
Re-pretraining a compact architecture informed by these sparsity patterns is a 
natural next step, discussed further in Section~\ref{outlook}.
Furthermore, post-training dynamic INT8 quantisation of the large model results in a minor reduction in motion classification accuracy (0.8301 to 0.8205), whilst reducing model size by approximately 75\% (1,390MB to 349MB on disk) and improving inference throughput by 53\% (12.2s to 5.7s per batch of 32). 
Quantisation evaluation was performed on an Azure NC16as T4 v3 instance (AMD EPYC 7V12, 16 vCPUs, 110 GiB RAM). Note that the GPU was not used for this benchmark, as PyTorch's quantisation requires CPU execution.

To provide a more conservative estimate of inference speed relevant to resource-constrained deployment (e.g., single-core embedded or onboard satellite compute, where multiple CPU cores may not be available), we additionally benchmarked FP32 and INT8 inference restricted to a single CPU core. This INT8 
speedup persisted, reducing inference time from 165.2s to 60.7s per batch 
(63.3\% faster); accuracy was consistent with the main multi-core results (83.0\% FP32, 82.1\% INT8).
Whilst preliminary, and a full quantisation/pruning analysis is outside the scope of this work, these results suggest that the model could be readily adapted to practical deployment in resource-constrained environments without substantial performance degradation. 
\section{Methods}
\label{sec:methods}
\subsection{Data Sources and Preparation}
\begin{table}[htbp]
\caption{Data modalities and their sources. For GMV GRIAL, sample counts per motion class are: I=Inertial, SA=Safe, SP=Spin, SU=Sun, T=Tumbling, Y=Yaw.\label{tab:pretraining_datasets}}
\vspace{5pt}
\resizebox{\columnwidth}{!}{%
\begin{tabular}{@{}lccc@{}}
\toprule
\textbf{Source} & \textbf{Notes} & \textbf{\textit{n}} & \textbf{Labels} \\
\midrule
MMT-9 & Real obs. & 227,697 & - \\
CASSANDRA & Simulated & 800 & 160 anom., 640 normal \\
GMV GRIAL & Simulated & 22,006 & I:2361, SA:3575, SP:2359, SU:1718, T:2500, Y:9493 \\
\bottomrule
\end{tabular}}
\end{table}
The MMT-9 light curve dataset was downloaded as text files from \url{https://www.sao.ru/lynx/karpov/satellites/} and processed with custom scripts using the BeautifulSoup Python library (v4.11.1). The light curves varied in length and were therefore resampled to 
$n_{\text{samples}} = 128$ time steps. Whilst the Perceiver architecture does 
not strictly require uniform input length, uniform resampling simplifies 
batching and ensures consistent positional encoding across samples. Visual inspection showed this length to not noticeably degrade light curves. Similarly, the anomaly detection (CASSANDRA) and motion prediction (GRIAL) fine-tuning datasets were processed to $(n_{\text{samples}}, 128, n_{\text{labels}})$, where there were two labels for anomaly detection and six for motion prediction. For details on dataset sizes, please see Table \ref{tab:pretraining_datasets}, which describes the datasets produced by the
CASSANDRA simulator (anomaly detection fine-tuning), and
GRIAL simulator (motion prediction fine-tuning).
\subsection{Model Architecture}

Our Perceiver-VAE model architecture is outlined in Equation \ref{diag:perceiver-vae}, and we briefly describe these steps below.

We implemented Perceiver as outlined in \cite{jaegle_perceiver_2021}, adapting the Pytorch implementation of \cite{wang2021perceiver}, including Fourier positional encoding (Figure \ref{fig:architecture-pretraining}, Equation \ref{diag:perceiver-vae}). This encoding maps temporal coordinates to higher-dimensional representations, enabling the attention mechanisms to capture sequential patterns in light curve data.

To include Variational Autoencoder functionality we computed the mean and log variance of the learned latent vector \textit{z}, using those values to sample the latent space probabilistically. This produces the reconstructed input, the Mean Squared Error loss (Equation \ref{eqn:recon_loss}), and KL Divergence, which constrains the latent space such that it is well ordered and stable (Equation \ref{eqn:kl_div}). By including the VAE sampling, (as opposed to standard autoencoding), we create a continuous, structured latent space that enables generation of novel data points through interpolation and sampling. For decoding, we use a standard feed forward network with the architecture shown in (Equation \ref{eqn:decoder}).
\begin{align}
\label{diag:perceiver-vae}
\scalebox{0.9}{ 
$\text{Perceiver-VAE} = \begin{cases}
\text{Encoder:} \\
\quad \text{Input} \xrightarrow{\text{Fourier Encoding}} \text{Encoded Position} \\
\quad \text{Latent Array} \xrightarrow{\text{Cross-Attention, FF}} \text{Updated Latents} \\
\quad \text{Updated Latents} \xrightarrow{\text{Self-Attention}} \text{Processed Latents} \\
\quad \text{Processed Latents} \xrightarrow{} \mu, \log\sigma^2 \\
\quad \left. z = \mu + \varepsilon \cdot \exp(0.5 \cdot \log\sigma^2) \right\} \text{\scriptsize reparam.} \\
\text{Decoder:} \\
\quad z \xrightarrow{} \text{Reconstruction}
\end{cases}$
}
\end{align}
Where
\begin{equation}
\label{eqn:decoder}
\text{Decoder} = \begin{cases}
\text{Linear}(d_{\text{latent}} \to 512) \\
\text{LayerNorm} \to \text{ReLU} \\
\text{Linear}(512 \to 1024) \\
\text{LayerNorm} \to \text{ReLU} \\
\text{Linear}(1024 \to d_{\text{output}})
\end{cases}
\end{equation}

We evaluate three model scales: small (64 latents, $d_z$=128, 
6.4M parameters), medium (128 latents, $d_z$=128, 13.2M parameters), 
and large (384 latents, $d_z$=512, 347.5M parameters), the latter 
representing the largest scale trainable within our computational budget. Magnitude pruning results for this model (Section~\ref{sec:pruning}) suggest 
substantial parameter redundancy at this scale, discussed further in 
Section~\ref{outlook}.
\subsection{Pre-training strategy}
\label{methods:pre-training_strategy}
We pre-trained our Perceiver-VAE model using a multi-task self-supervised learning approach.
For this we incorporated three objectives: reconstruction, masking, and forecasting.
The reconstruction task trained the model to encode and decode whole light curves, establishing the foundations for latent space representation. 
At the same time, the masking task required the model to predict randomly masked segments of input sequences, encouraging our model to learn contextual relationships within the light curve.
Finally, the forecasting task trained the model to predict future sequence values, promoting the capture of temporal dynamics and enabling forecasting at inference time. 

Each sample is presented to the model once per batch in a single forward pass. 
The reconstruction, masking, and forecasting outputs are computed simultaneously 
within this pass, with each task operating on a different view of the same input 
(full sequence, randomly masked sequence, and future-masked sequence respectively). 
The KL divergence is therefore computed once per sample from the single latent 
sample, and the effective batch size equals the nominal batch size throughout 
pre-training.
We combined these learning objectives by summing individual mean squared error loss components for each task, combined with a KL divergence term (weight $\alpha$ = 0.001) to regularise the latent space (Equation \ref{eqn:total_loss}). This combined optimisation enabled the model to learn rich representations of light curve characteristics whilst maintaining a structured latent space suitable for generation and interpolation.
\begin{align}
    \mathcal{L}_{\text{total}} &= \mathcal{L}_{\text{recon}} + \alpha \mathcal{L}_{\text{KL}} + \mathcal{L}_{\text{mask}} + \mathcal{L}_{\text{forecast}}, \label{eqn:total_loss}\\
    \text{where:} \nonumber \\
    \mathcal{L}_{\text{recon}} &= \frac{1}{N} \sum_{i=1}^{N} (x_i - \hat{x}_i)^2, \label{eqn:recon_loss}\\
    \mathcal{L}_{\text{mask}} &= \frac{1}{N_m} \sum_{i \in M} (x_i - \hat{x}_i^m)^2, \label{eqn:mask_loss}\\
    \mathcal{L}_{\text{forecast}} &= \frac{1}{N_f} \sum_{i \in F} (x_i - \hat{x}_i^f)^2.
    \label{eqn:forecast_loss}\\
    \mathcal{L}_{\text{KL}} &= -\frac{1}{2} \sum_{j=1}^{d} (1 + \log\sigma_j^2 - \mu_j^2 - \sigma_j^2). 
    \label{eqn:kl_div}
\end{align}
The total loss (Equation \ref{eqn:total_loss}) combines four components: The reconstruction loss ($\mathcal{L}_{{\text{recon}}}$, Equation \ref{eqn:recon_loss}) measures the Mean Square Error (MSE) between the original inputs $x_i$ and reconstructed outputs $\hat{x}_i$ across all $N$ data points. 

The masking loss ($\mathcal{L}_{{\text{mask}}}$, Equation \ref{eqn:mask_loss}) calculates MSE on a subset of masked inputs ($M$), where $\hat{x}_i^m$ represents reconstructions of deliberately masked elements, encouraging the model to infer missing data from context. 

The forecasting loss ($\mathcal{L}_{{\text{forecast}}}$, Equation \ref{eqn:forecast_loss}) computes MSE on future time steps ($F$), where $\hat{x}_i^f$ represents predicted values for future time points, training the model to extrapolate temporal patterns.

The KL divergence loss ($\mathcal{L}_{{\text{KL}}}$, Equation \ref{eqn:kl_div}) regularises the latent space by ensuring the learned distribution (parameterised by mean $\mu_j$ and variance $\sigma_j^2$ across $d$ latent dimensions) approximates a standard normal distribution, weighted by hyperparameter $\alpha$. 

Pre-training hyperparameters were selected via Bayesian optimisation using Optuna \citep{akiba_optuna_2019}, running 25 trials of 15 epochs each per model scale. 
The optimal learning rates were $1.41 \times 10^{-3}$, $5.56 \times 10^{-3}$, and 
$5.38 \times 10^{-4}$ for small, medium, and large respectively, with batch sizes 
of 16, 32, and 16. All models were trained with gradient clipping at 0.5 \citep{pascanu_difficulty_2013} 
to prevent exploding gradients, which we observed empirically during the early training 
phase where the KL divergence term increases rapidly before stabilising 
(Figure \ref{fig:train_loss_curves}). A learning rate scheduler reducing the learning rate by a factor of 0.5 was used when validation loss plateaued for 5 consecutive epochs.

To assess potential confounds in reconstruction error due to observing conditions, we computed the Spearman rank correlation \citep{virtanen_scipy_2020} between per-curve reconstruction error and mean observing distance as a proxy for signal-to-noise ratio, across the held-out test set. Since Spearman's $\rho$ operates on ranks rather than raw values, variance explained was quantified as R$^2$ from ordinary least squares regression applied to rank-transformed variables (i.e., the ordinal positions of reconstruction errors and distances when sorted from lowest to highest), preserving consistency with the monotonic assumptions of the rank correlation.

All pre-training was conducted on the Baskerville HPC cluster, using single A40 and A100 GPUs.
\subsection{Fine-tuning}
\label{sec:fine-tuning_methods}

To fine-tune our pre-trained Perceiver-VAE, we freeze the encoder weights and train a lightweight classification head on top of the learned latent representations \citep{devlin_bert_2019}. For a given input light curve, the frozen encoder produces a mean latent vector $\mu$, which is passed directly to the classification head. The optimisation objective is therefore simply the categorical cross-entropy loss:

\begin{equation}
    \mathcal{L}_{\text{fine-tune}} = \mathcal{L}_{\text{class}} = -\frac{1}{m} \sum_{i=1}^{m} \sum_{c=1}^{C} y_{ic} \log(\hat{y}_{ic}), \label{eqn:fine-tune_loss}
\end{equation}

across $m$ samples and $C$ classes, where $y_{ic}$ are the true labels and $\hat{y}_{ic}$ the predicted probabilities. This approach follows standard practice for fine-tuning foundation models, using the pre-trained encoder purely as a feature extractor.

The classification head is a two-layer MLP: a linear projection to a hidden dimension, followed by ReLU activation, dropout regularisation ($p=0.3$), and a final linear projection to the number of output classes.

We trained using the Adam optimiser with a learning rate of $1\times10^{-3}$ for 
all fine-tuning conditions. A batch size of 32 was used across both tasks, with 
maximum epochs set per task (50 for anomaly detection, 150 for motion prediction), using early stopping with a patience of 50 epochs based on validation accuracy. 
To address class imbalance in the motion prediction task, (Table \ref{tab:motion_classification_metrics}), we applied inverse-frequency class weighting to the cross-entropy loss. The best checkpoint was selected by validation ROC AUC.

For motion prediction in Tables \ref{tab:baseline_comparison}-\ref{tab:motion_classification_metrics}, we additionally evaluate end-to-end fine-tuning, 
in which the encoder weights are unfrozen and updated alongside the 
classification head throughout training, using the same optimiser 
settings and cross-entropy objective. This allows the encoder 
representations to adapt to the downstream task rather than acting 
purely as a fixed feature extractor. For anomaly detection (Table \ref{tab:anomaly_comparison}), end-to-end fine-tuning did not improve performance, consistent with the frozen 
encoder already achieving strong performance in the low-label regime.
To evaluate the contribution of pre-training, we additionally train 
classification heads on randomly initialised Perceiver encoders of 
equivalent scale, with weights drawn from the default PyTorch 
initialisation and held frozen throughout fine-tuning. These 
random encoder (r.e.) variants are otherwise identical to the 
pre-trained models, isolating the effect of pre-training on 
downstream performance.

\subsubsection{Anomaly Detection Fine-tuning}
For anomaly detection, the classification head produces a binary output distinguishing anomalous from normal light curves. We evaluated on a stratified held-out test set comprising 20\% of the CASSANDRA dataset, reporting both accuracy and ROC AUC.

To evaluate label efficiency, we additionally conducted a $k$-shot 
study comparing the pre-trained Perceiver against the CNN baseline. For each value of $k \in \{2, 5, 10, 25\}$, $k$ labelled examples per class were sampled from the training pool using independent random seeds across 20 independent runs. Each run trained the classification head (or CNN) for 50 epochs using the Adam optimiser with a learning rate of $1\times10^{-3}$ and batch size of 32, and was evaluated on the same stratified held-out test set used in the main experiments. 

\subsubsection{Motion Prediction Fine-tuning}
For motion prediction, the classification head produces a six-class output corresponding to the grouped motion classes described in Table \ref{tab:motion_classification_metrics}. We evaluated on a stratified 10\% held-out set (n = 2201), with the remaining data split 80/20 into training and validation sets, reporting per-class precision, recall, F1-score, and ROC AUC.

\subsection{Synthetic Data Generation}
\label{sec:syndata_methods}
To explore the latent space and synthetic data generation capabilities of our model, we implemented a neighbourhood sampling approach.
For synthetic data generation, our objective is to sample the latent space in a way that produces meaningful variations of a reference curve whilst preserving its class identity. 

Here we use the fine-tuned motion prediction classifier to generate a set of high confidence 'query' curves. The algorithm first identifies some reference curves above a given confidence threshold (0.95) for a requested motion class e.g., tumbling. Then, it uses these to index the learned latent space of the model, and samples around that latent space by adding Gaussian noise to the highest activating neurons.
For visual clarity in Figure~\ref{fig:ati:syndata}, Gaussian smoothing 
($\sigma = 2.0$) is applied to the displayed synthetic curves to aid 
inspection of global morphological features. 

To quantitatively validate generated synthetic light curves, we fitted a Gaussian Process (GP) to the real light curve distribution for each motion class and evaluated whether synthetic curves fell within the learned distributional bounds.

For each class, the held-out test split was partitioned into three disjoint subsets using true class labels: anchor curves (used to seed synthetic generation), GP-fit curves (used to fit the GP), and real baseline curves (scored against the fitted GP as a reference for natural sampling variation). This ensures that neither the generation process nor the classifier confidence screening influences the anchors against which synthetic curves are evaluated.

A GP was fitted to the GP-fit curves using an RBF kernel with additive white noise:
$k(t,t') = \text{RBF}(l{=}0.2) + \text{WhiteKernel}(\sigma^2{=}0.01)$,
with kernel hyperparameters refined via three restarts of marginal likelihood optimisation and outputs normalised prior to fitting. This provides a non-parametric estimate of the mean and standard deviation of the real light curve distribution at each of the 128 resampled time steps.

Each synthetic curve was scored by computing the fraction of time steps for which the absolute deviation from the GP posterior mean fell within two standard deviations:
\begin{equation}
    s = \frac{1}{128} \sum_{i=1}^{128} 
    \mathbf{1}\!\left[|\hat{y}_i - \mu_i| < 2\sigma_i\right],
    \label{eqn:gp_score}
\end{equation}
where $\mu_i$ and $\sigma_i$ are the GP posterior mean and standard deviation at time step $i$, and $\hat{y}_i$ is the synthetic curve value. Real baseline curves were scored against the same fitted GP using the same metric, providing a reference score for genuinely real curves under the GP model. The class-level score is the mean of $s$ across all synthetic curves for that class. A high score indicates synthetic curves are statistically consistent with real observations under the GP model.

The GP validation 
(Section~\ref{sec:synthetic_data_generation}, Figure~\ref{fig:gp_validation}) is 
performed on unsmoothed outputs, confirming that quantitative synthetic curve 
quality is maintained without smoothing.

\subsection{Model Compression}
\label{sec:compression_methods}
Magnitude pruning~\citep{han_learning_2015} was applied separately to the 
Perceiver layers and the VAE bottleneck (mean and log-variance projections), 
sweeping sparsity from 10\% to 90\%. Following pruning, batch normalisation 
statistics were recalibrated on 500 training samples prior to evaluation on 
the held-out test set.
Post-training dynamic quantisation was applied to the full model, converting 
all linear layers to INT8 precision using PyTorch's \texttt{quantize\_dynamic}. 
No calibration data was required.
\section{Outlook}
\label{outlook}
Our work is the first to demonstrate that a VAE architecture can effectively learn meaningful representations from light curve data and be fine-tuned for multiple downstream tasks including detecting anomalous space object behaviour and motion prediction.
Our architecture is designed to scale efficiently to large datasets whilst maintaining computational tractability, which is critical considering the rapidly growing amount of Space Object data.
Here, we developed a two-stage approach, first through unsupervised pre-training, whereby the model learns rich features which aid reconstruction of real light curves.
Then we performed supervised fine-tuning, which leverages those features for both anomaly classification, reaching 85\% accuracy with a 0.92 ROC AUC score, and motion prediction, achieving 82.7\% test accuracy, with 0.95 ROC AUC scores.
Additionally, we demonstrate that the same architectural framework and approach can simultaneously achieve multiple space object behavioural analysis goals (i.e., anomaly detection, attitude mode classification, and synthetic data generation), provided that the representations learned during pre-training are diverse and rich, and that these are coupled with high-fidelity fine-tuning simulation data.
Although in this study we identify several predicted anomalous light curve modes, some present clearly anomalous behaviour (e.g., rapid tumbling), whilst for others the underlying behaviour is less clear.
A further direction is per-object anomaly characterisation: for 
well-represented objects in the pre-training data, fitting a Gaussian 
Process to the distribution of typical light curves for a specific 
object and comparing high reconstruction error curves against this 
baseline would allow more rigorous validation of whether these curves 
represent genuinely atypical behaviour for that object specifically.
Systematic cataloguing and analysis of simulation conditions that lead to these various curve morphologies would allow us to verify whether particular erroneous space object dynamics produce these curves (e.g., in Figure \ref{fig:ati:fine-tuning_predictions}C, F). A further valuable direction is a systematic pre-training scaling study, training models on increasingly large subsets of the MMT-9 dataset to characterise how pre-training data volume impacts performance. This is of particular relevance to the SDA context, where pre-training data may grow substantially as sensor networks expand, but detailed labels remain sparse.

\bigskip
Untangling morphological similarities in observations is another direction for future work. Possible mitigations include additional discriminative features that better separate these morphologically similar classes, hierarchical classification that first groups slowly-varying motions (INERTIAL, YAW) before a finer-grained second stage, or targeted data augmentation that increases the diversity of INERTIAL examples relative to YAW. A physical mechanism may also contribute to this ambiguity. For example, where a space object's geometry and surface materials are approximately symmetric about its rotation axis, attitude changes would produce little variation in reflected flux, potentially rendering YAW and INERTIAL motions sometimes near-identical in a single light curve.
If YAW and INERTIAL prove intrinsically difficult to separate from single light curves, incorporating sequences of observations of the same object over multiple orbits may help disentangle the two, since YAW's orbit-locked corrections should become apparent over a longer time horizon where a single snapshot is not sufficient.
\bigskip

From this work, several key directions for further development emerge. Firstly, development of robust quantitative benchmarks for AI synthetic data. Whilst our approach encourages plausibility in the generated curves through reference-based sampling, systematic evaluation frameworks are needed to validate produced synthetic data beyond visual inspection/sense checks.
Whilst we provide GP-based validation of synthetic light curve 
quality in this work, further quantitative evaluation metrics 
such as KL divergence, power spectral density comparisons, or 
Earth mover distance between real and synthetic distributions 
remain valuable directions for future work.
For example, these frameworks could assess statistical distribution matching between real and AI generated datasets, and evaluate like-for-like comparisons in downstream tasks trained on real data, numerically simulated data, and/or data generated through our approach (i.e., generative AI data). Finally, physics-informed neural networks (PINNs) could be used to quantify physical plausibility.
Briefly, these networks combine a data loss term with a physics loss term. 
For example, for light curves this might be a reflectivity model as outlined in \cite{matsushita_light_2019}. At inference time, a well-trained PINN will give a good indicator as to whether a synthetically generated light curve is obeying the underlying physics due to the value of the physics loss term.

A key limitation of the current work is the simulation-to-real gap: both  fine-tuning datasets (CASSANDRA for anomaly detection and GRIAL for motion  prediction) are simulated, whilst pre-training data and any operational inference data are real observations. 
This means reported downstream accuracies reflect performance on simulated distributions, and should be interpreted as an upper bound pending validation against real labelled examples as these become available.
Whilst the pre-trained model demonstrates some unsupervised anomaly discrimination on the CASSANDRA dataset (ROC AUC 0.71 without any labels), the domain gap between real and simulated curves limits threshold-based detection, motivating fine-tuning as a necessary bridging step. Closing this gap will require either real labelled anomaly data, 
which is currently scarce due to national security and commercial sensitivities, or domain adaptation techniques that better align simulated and real light curve distributions. Contrastive approaches using paired real and simulated observations of the same object could be a promising direction, as could physics-informed data augmentation to make simulated curves more representative of real observational 
conditions.

Finally, magnitude pruning experiments (Section~\ref{sec:pruning}) suggest 
the large model contains substantial redundancy in its Perceiver layers, motivating 
future work on more parameter-efficient pre-training strategies.
Similarly, the negligible accuracy cost of INT8 quantisation suggests that 
deployment of large-scale space object characterisation models in 
resource-constrained operational environments is feasible.
\bigskip

Another key direction is privacy and security considerations. As generative AI (and therefore also synthetic data) becomes more prevalent in SDA, privacy/security need to be considered to ensure that if a model is pre-trained on sensitive light curves that they cannot be extracted from the resulting fine-tuned model at inference time.
Additionally, the quality of a self-supervised model like ours is dependent on the quality of both the pre-training and fine-tuning data. If models such as these are to be integrated into critical Space Domain/Situational Awareness pipelines they become an attractive target for adversarial attack, particularly through data poisoning.
For this reason, ensuring the integrity of data provenance is highly important, and blockchain approaches could be used here, treating the dataset as the transaction to be verified on the public ledgers, e.g., through the use of Trustchain \citep{hobson_trustchain_2023}.

\bigskip

A substantial challenge direction for further development is multi-modal fusion, i.e., integrating the light curve analysis presented here with other observation types (e.g., hyperspectral data, orbits, radar etc.).
Indeed, part of the decision to use a Perceiver based architecture was to readily enable extension to other data modalities, as it is well suited to this through its cross-attention and latent bottleneck design, which could enable multiple modalities to be reduced to a common set of embeddings. A natural extension of future work is to exploit the Perceiver's 
ability to handle variable-length inputs directly, removing the 
uniform resampling constraint applied here and allowing the model 
to process light curves at their native sampling rates.
However, there are substantial technical challenges in multimodal fusion, including developing an appropriate fusion strategy at different levels of the pre-training/fine-tuning pipeline. One such method to fuse information from other sensors with light curves could include implementation of contrastive learning using paired observations. For example, positive pairs (same space object at different dates) and negative pairs (different objects on the same date) would help develop more discriminative features for space object characterisation.
Additionally incorporating metadata prediction tasks as part of the self-supervised framework could help capture characteristic signatures of specific satellites, or debris. For example, a fine-tuned metadata prediction FM could generate a likely light curve given an orbital eccentricity, which could help unpick the relationships between satellite structure/behaviour and orbital path.

Our initial framework provides promising results, and as the amount of data available increases, a basis for further analysis. We suggest Foundation Models, in particular those that exploit multimodal datasets, could become powerful tools for automated space object monitoring and space situational awareness.
As orbital populations continue to grow, automated approaches such as ours will become increasingly crucial for maintaining space safety \& sustainability.

\section*{Funding}
This work was supported by the UKSA International Bilateral Fund project: AI for Space Operations, Safety
and Sustainability (AI4S3), Project Number: UKSAG23 0042-081.

\section*{Conflict of interest}

The authors declare no potential conflict of interests.
 
\section*{Author contributions}

CRediT Classification:

\textbf{Conceptualization:} IG, MV, VN
\textbf{Data Curation:} IG, AC, JF, DRR
\textbf{Formal Analysis:} IG
\textbf{Funding Acquisition:} PM, MV, VN
\textbf{Investigation:} IG
\textbf{Methodology:} IG, MV, VN
\textbf{Project Administration:} IG, MV, VN
\textbf{Software:} IG, AC
\textbf{Supervision:} VN
\textbf{Writing – Original Draft Preparation:} IG
\textbf{Writing – Review \& Editing:} IG, VN

\section*{Data Availability Statement}
Research data are not shared.
\bibliography{main}  

\section*{Supplementary}
\subsection*{Supplementary: Reconstruction Error vs.\ Observing Distance}
Figure~\ref{fig:supp:recon_distance} shows the relationship between per-curve reconstruction error and mean observing distance across the held-out test set ($n = 24{,}995$), corresponding to the analysis described in Section~\ref{sec:pre-flagged_anomalies_analysis}.
For details on implementation, please refer to Section \ref{methods:pre-training_strategy}.

\begin{suppfigure}[h]
    \centering
    \includegraphics[width=0.8\columnwidth]{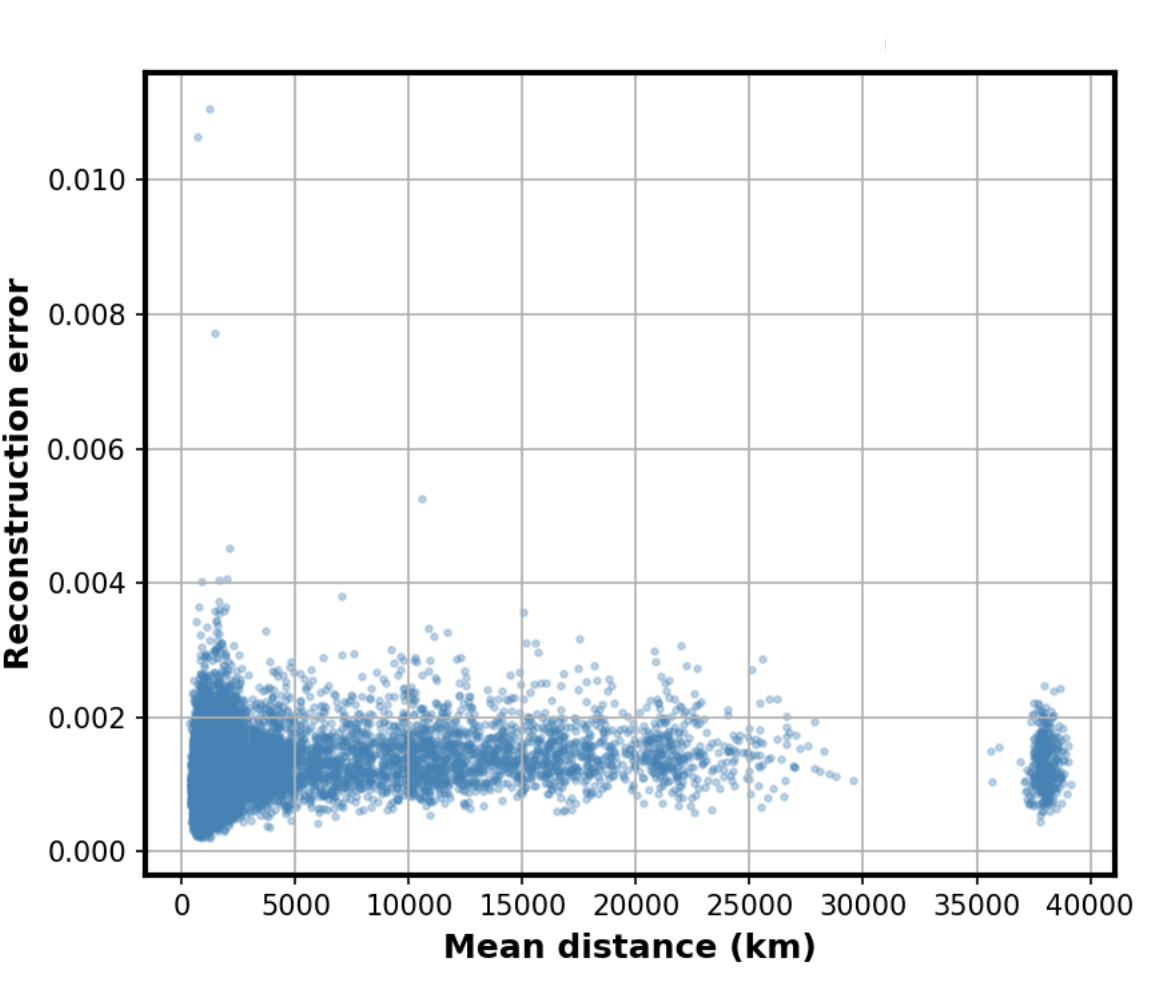}
    \caption{Reconstruction error as a function of mean observing distance (km) across the MMT-9 held-out test set. Each point represents one light curve. Spearman $\rho = 0.34$ ($p < 0.001$), with approximately 11\% of variance in reconstruction error explained by observing distance (R$^2$ from Ordinary Least Squares on rank-transformed variables).}
    \label{fig:supp:recon_distance}
\end{suppfigure}

\subsection*{Supplementary: Motion Classification Confusion Matrix}
Figure~\ref{fig:supp:confusion_matrix} shows the row-normalised confusion matrix for the motion classification model evaluated on the held-out test set, corresponding to the analysis described in Section~\ref{sec:fine-tuning_motion}.

\begin{suppfigure}[h]
    \centering
    \includegraphics[width=0.8\columnwidth]{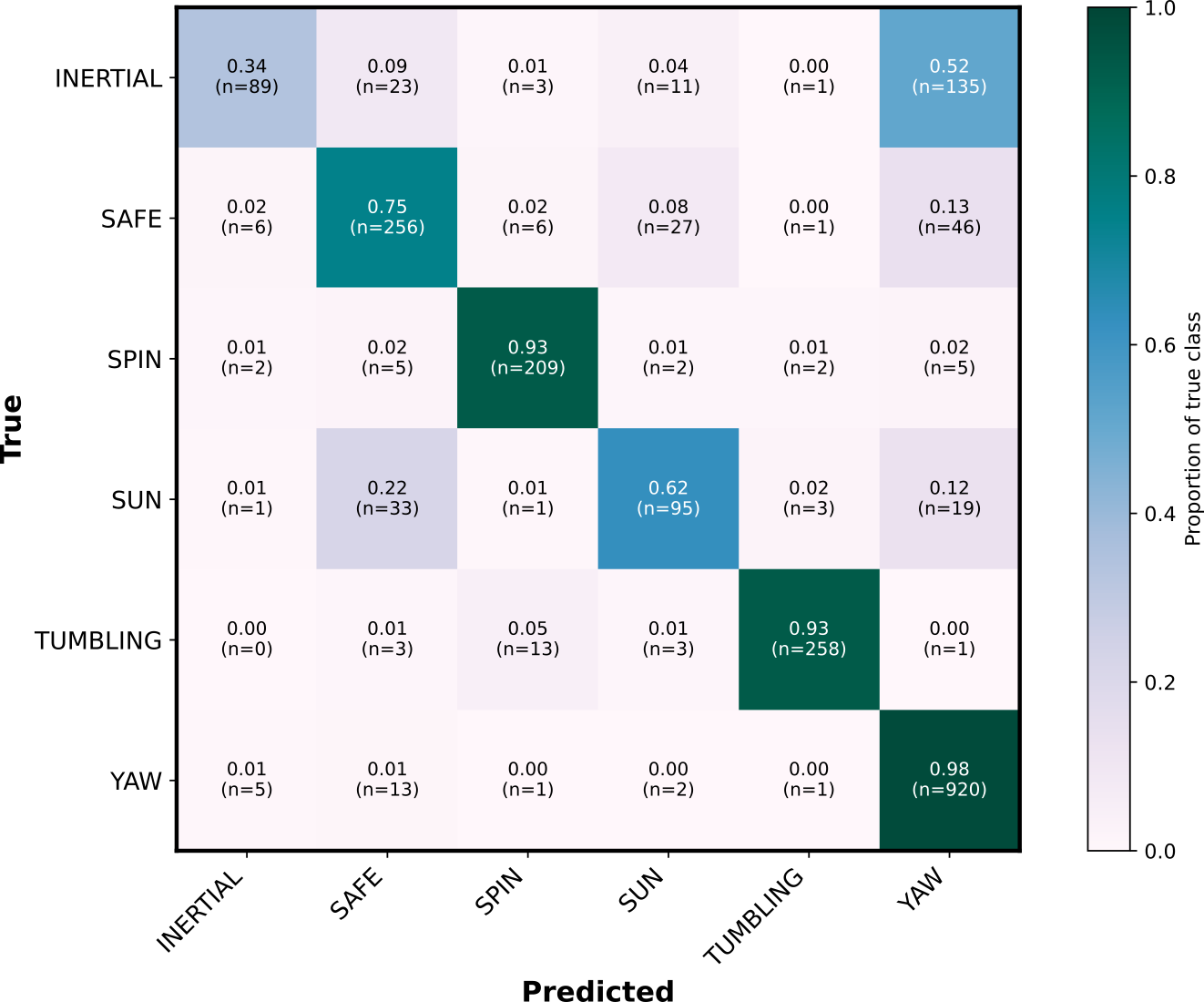}
    \caption{Confusion matrix for the motion classification model on the held-out test set ($n = 2{,}201$). Each cell shows the proportion of true-class samples predicted as each class, with raw counts in parentheses. INERTIAL shows high precision but low recall, with the majority of misclassifications concentrated in the YAW class, consistent with the morphological similarity between these two motion types discussed in Section~\ref{sec:fine-tuning_motion}.}
    \label{fig:supp:confusion_matrix}
\end{suppfigure}

\subsection*{Baseline Model Architectures}
All baseline models take a light curve of length 128 as input and are 
trained from scratch using the Adam optimiser with a learning rate of 
$1\times10^{-3}$, batch size of 32, and dropout of $p=0.3$ throughout.
Baselines were trained and evaluated on identical train/validation/test splits 
and the same preprocessing pipeline as the pre-trained model (Section~\ref{sec:fine-tuning_methods}), 
using the same random seed. As with the pre-trained model's classification 
heads, baseline hyperparameters were not subject to optimisation.

\textbf{RNN:} A two-layer recurrent neural network with \texttt{tanh} 
activation and hidden size 64, following a standard stacked configuration 
to capture both short and long-range sequential dependencies. The final 
hidden state is passed to a linear output projection.

\textbf{LSTM:} A two-layer Long Short-Term Memory network with hidden 
size 64, again using a stacked configuration for multi-scale temporal 
modelling. The final hidden state is passed to a linear output projection.

\textbf{CNN:} Three successive 1D convolutional blocks with progressively 
increasing channel depths (32, 64, 128) and kernel sizes of 7, 5, and 3 
respectively, to capture hierarchical temporal features at decreasing 
scales. Each block is followed by ReLU activation and max pooling. 
Global average pooling precedes a linear output projection.

\textbf{Transformer:} A two-layer Transformer encoder with model dimension 
$d_{\text{model}}$=64, 4 attention heads, and feedforward dimension 256, 
following the standard 1:4 model-to-feedforward scaling convention 
\citep{vaswani_attention_2017}. Input timesteps are projected to 
$d_{\text{model}}$ before encoding, and global average pooling over 
the sequence precedes a linear output projection.

\end{document}